\documentclass[10pt,journal,compsoc]{IEEEtran}

\usepackage[table,x11names]{xcolor}

\usepackage{amsmath,amsfonts}
\usepackage{array}

\definecolor{cyan}{RGB}{0,255,255}
\usepackage{textcomp}
\usepackage{stfloats}
\usepackage{url}
\usepackage{verbatim}
\usepackage{cite}
\usepackage{enumerate}
\usepackage{times}
\usepackage{graphicx}
\usepackage{amssymb}
\usepackage{paralist,bbding,pifont}
\usepackage{multirow}
\usepackage{amsfonts}
\usepackage{booktabs}
\usepackage[pagebackref=true,colorlinks,bookmarks=false,citecolor=blue,linkcolor=blue,urlcolor=blue]{hyperref}
\usepackage{cleveref}
\usepackage{caption}
\usepackage[ruled,linesnumbered,boxed]{algorithm2e}
\usepackage{ragged2e}
\usepackage{setspace}
\usepackage{fancyhdr}
\usepackage{xspace}
\usepackage{color}
\usepackage{multirow}
\usepackage{ifthen}
\usepackage{tcolorbox}
\renewcommand{\baselinestretch}{1}

\newcommand{\mb}[1]{\mathbf{#1}}

\long\def\ignorethis#1{}

\definecolor{gray}{rgb}{0.35,0.35,0.35}
\definecolor{MyBlue}{rgb}{0,0.2,0.8}
\definecolor{MyRed}{rgb}{0.8,0.2,0}
\definecolor{MyGreen}{rgb}{0.0,0.5,0.1}
\definecolor{MyGray}{rgb}{0.4,0.4,0.4}

\newlength\paramargin
\newlength\figmargin
\newlength\subfigmargin
\newlength\secmargin
\newlength\subsecmargin
\newlength\tabmargin
\newlength\eqmargin

\usepackage{array}
\newcolumntype{L}[1]{>{\raggedright\let\newline\\\arraybackslash\hspace{0pt}}m{#1}}
\newcolumntype{C}[1]{>{\centering\let\newline\\\arraybackslash\hspace{0pt}}m{#1}}
\newcolumntype{R}[1]{>{\raggedleft\let\newline\\\arraybackslash\hspace{0pt}}m{#1}}

\definecolor{grey}{rgb}{0.5, 0.5, 0.5}
\def\ie{i.e.,~}
\def\eg{e.g.,~}
\def\etc{etc}

\def\vs{vs.~}

\newcommand{\subsecref}[1]{Section~\ref{subsec:#1}}
\newcommand{\figref}[1]{Fig.~\ref{fig:#1}}
\newcommand{\tabref}[1]{Table~\ref{tab:#1}}
\newcommand{\eqnref}[1]{Eq.~\eqref{eq:#1}}

\newcommand{\algref}[1]{Algorithm~\ref{#1}}

\newcommand{\Paragraph}[1]{\noindent\textbf{#1}}

\begin{document}

% \title{Towards Understanding Cross-attention Alignment in Stable Diffusion for Localized Image Editing}
%\title{UnifyEdit: Balancing Fidelity and Editability in Tuning-Free Text-Based Image Editing 
%\\
%via Adaptive Latent Optimization}
%
\title{Tuning-Free Image Editing
with Fidelity and Editability 
via Unified Latent Diffusion Model
}
%\title{Tuning-Free Text-Based Image Editing
%with Fidelity and Editability 
%via Unified Latent Diffusion Model
%}
%\title{UnifyEdit: Unified Diffusion Latent Optimization for Balancing Fidelity and Editability in Tuning-Free Text-Based Image Editing}
\author{Qi Mao, Lan Chen, Yuchao Gu, Mike Zheng Shou, Ming-Hsuan Yang
\thanks{Qi Mao and Lan Chen are with the State Key Laboratory of Media Convergence and Communication, Communication University of China. (E-mail: qimao@cuc.edu.cn, chenlaneva@mails.cuc.edu.cn). \protect\\
Yuchao Gu and Mike Zheng Shou are with Show Lab, National University of Singapore. (E-mail: yuchaogu@u.nus.edu, mikeshou@nus.edu.sg).
\protect\\
Ming-Hsuan Yang is with the University of California at Merced and Yonsei University. (E-mail: mhyang@ucmerced.edu).\\
(Corresponding Author: Qi Mao)}
}

%\author{Qi Mao~\IEEEmembership{Member,~IEEE,} Lan Chen, Yuchao Gu, Mike Zheng Shou, Ming-Hsuan Yang~\IEEEmembership{Fellow,~IEEE}}

        % <-this % stops a space

\markboth{Under Review}%
{}

% \maketitle
\IEEEtitleabstractindextext{%
\begin{abstract}
\justifying
Balancing fidelity and editability is essential in text-based image editing (TIE), where failures commonly lead to over- or under-editing issues. 
Existing methods typically rely on attention injections for structure preservation and leverage the inherent text alignment capabilities of pre-trained text-to-image (T2I) models for editability, but they lack explicit and unified mechanisms to properly balance these two objectives.
In this work, we introduce \emph{UnifyEdit}, a tuning-free method that performs \emph{diffusion latent optimization} to enable a balanced integration of fidelity and editability within a \emph{unified} framework.
Unlike direct attention injections, we develop two attention-based constraints: a self-attention (SA) preservation constraint for structural fidelity, and a cross-attention (CA) alignment constraint to enhance text alignment for improved editability.
However, simultaneously applying both constraints can lead to gradient conflicts, where the dominance of one constraint results in over- or under-editing. 
To address this challenge, we introduce an adaptive time-step scheduler that dynamically adjusts the influence of these constraints, guiding the diffusion latent toward an optimal balance.
Extensive quantitative and qualitative experiments validate the effectiveness of our approach, demonstrating its superiority in achieving a robust balance between structure preservation and text alignment across various editing tasks, outperforming other state-of-the-art methods.
%
%MH: I add this line. You can create a dummy site on GitHub but do not put the code there. 
The source code will be available at \href{https://github.com/CUC-MIPG/UnifyEdit}{https://github.com/CUC-MIPG/UnifyEdit}.
\end{abstract}

}    
\maketitle
\IEEEdisplaynontitleabstractindextext

\section{Introduction}
\label{sec:intro}
\IEEEPARstart{N}{atural} language is one of the most intuitive and effective ways for people to express their thoughts.
Recent advancements in large-scale text-to-image (T2I) diffusion models~\cite{ramesh2022hierarchical,rombach2022high,saharia2022photorealistic} have successfully bridged the gap between textual and visual modalities, 
facilitating the generation of high-quality images from free-form text prompts. 
However, in addition to creating images from scratch, there is an increasing need to modify existing images based on textual descriptions. 
This has led to the emergence of text-based image editing (TIE) \cite{brooks2023instructpix2pix, zhang2023magicbrush,kawar2023imagic,zhang2023sine,hertz2022prompt,tumanyan2023plug,parmar2023zero,mokady2023null,cao_2023_masactrl,couairon2022diffedit,magedit, avrahami2023blended,qiao2024baret,li2024source,titov2024guide}, which aims to manipulate input images according to given text prompts while preserving the integrity of other content. 
Over the past years, diffusion models for TIE have been extensively developed, categorized into:
instruction-based training~\cite{brooks2023instructpix2pix,zhang2023magicbrush}, fine-tuning~\cite{kawar2023imagic,zhang2023sine}, and tuning-free~\cite{hertz2022prompt,tumanyan2023plug,parmar2023zero,mokady2023null,couairon2022diffedit,cao_2023_masactrl,magedit} methods.
This work focuses on tuning-free editing approaches, which adapt existing T2I models for manipulations without extensive retraining or fine-tuning.

\begin{figure*}[!t]
    \centering
    \includegraphics[width=1.0\linewidth]{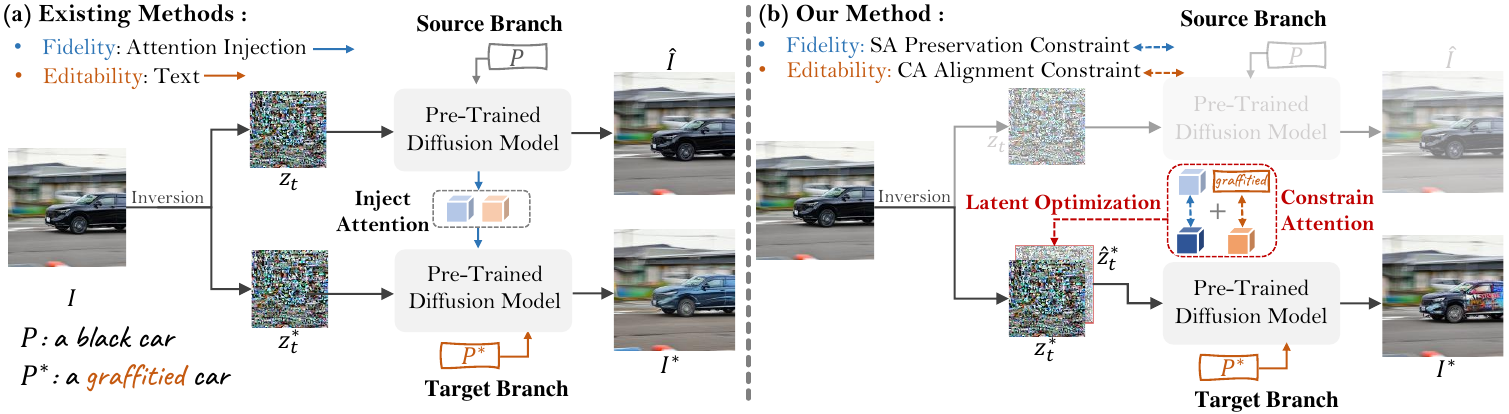} 
    \vspace{-5 mm}
    \caption{
    \textbf{UnifyEdit \vs dual-branch editing paradigm.}
    (a) The typical dual-branch editing paradigm consists of source and target branches, using \emph{attention injection} to maintain fidelity while relying on the \emph{text prompt} to achieve editability.
    (b) In contrast, our method explicitly models the fidelity and editability using two \emph{attention-based constraints} and performs \emph{latent optimization} within a unified framework, facilitating an adaptive balance across various editing types.
    }
    \label{fig:difference}
    \vspace{-4 mm}
\end{figure*}

Two critical concepts in TIE, distinct from T2I generation, are ``fidelity'' and ``editability''. 
\emph{{Fidelity}} concerns preserving the original image's content in areas that are not intended to be changed.
\emph{{Editability}} refers to the effectiveness of an editing method in making the desired changes specified by the text prompt. 
In the realm of diffusion models for TIE, a dual-branch editing paradigm such as P2P~\cite{hertz2022prompt} is commonly adopted, as demonstrated in~\figref{difference}(a).
This approach involves a \emph{source branch} that reconstructs the original image based on the source prompt and a \emph{target branch} that generates the target image guided by the target prompt. 
Within this framework, fidelity is achieved through shared inverted noise latents~\cite{song2020denoising, li2024source,ju2023direct} and structural information provided by the source branch~\cite{hertz2022prompt,tumanyan2023plug,cao_2023_masactrl,liu2024towards,qiao2024baret}. 
Meanwhile, editability originates from the inherent ability of T2I models to align target text descriptions with visual outputs in the target branch.

\begin{figure*}[!t]
    \centering
    \includegraphics[width=1.0\linewidth]{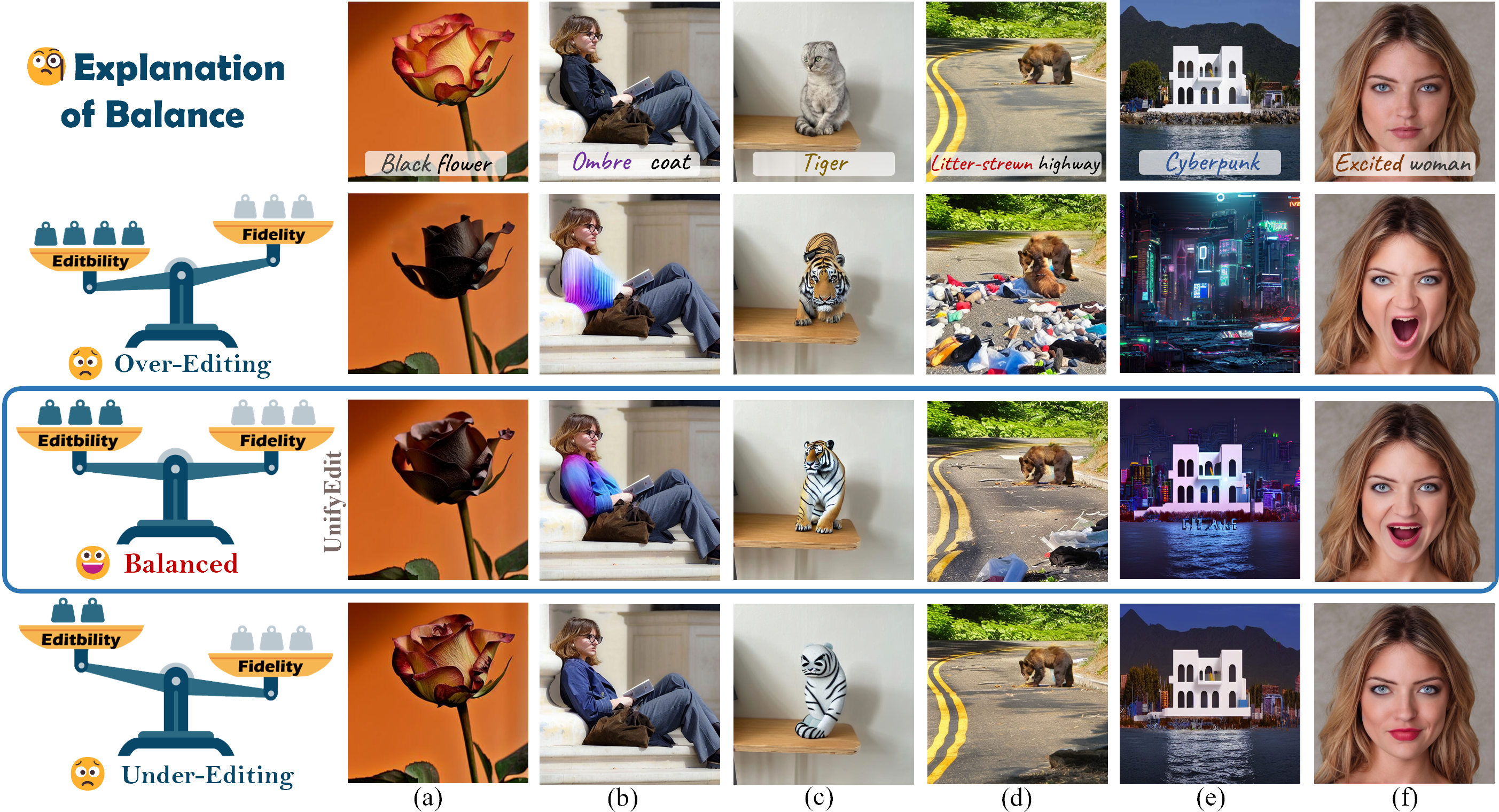} 
    \vspace{-6 mm}
    \caption{
    \textbf{Illustration of balancing fidelity and editability.}
    We demonstrate examples of over-, balanced, and under-editing across six types of edits: 
    (a) color change, (b) texture modification (c) object replacement (d) background editing, (e) global style transfer, and (f) human face attribute editing.
 Over-editing occurs when excessive changes distort the original image, while under-editing results in changes too subtle to meet the text prompt's requirements. 
In contrast, our \emph{\textbf{UnifyEdit}}  balances fidelity and editability within a unified framework, ensuring edits align with the text prompt while preserving the essential integrity.
    }
    \label{fig:balance}
\end{figure*}

The fundamental challenge in achieving this balance stems from the varying trade-offs required by different types of edits.
% As illustrated in \figref{balance}, 
For instance, color edits (\eg \figref{balance}(a)) demand a high degree of structural consistency to maintain the integrity of the original image. 
%\tabref{editing_type} and 
In contrast, object replacements (\eg \figref{balance}(c)) allow for greater editability, requiring only that the pose of the original elements be preserved.
As such, a poor balance can lead to two undesirable issues in \figref{balance}:
\begin{compactitem}
 \item {Over-editing} (editability $>$ fidelity): This occurs when the editing method makes excessive changes, prioritizing the text prompt over the original image's content. 
 For example, in the second row of
\figref{balance}(c), although the tiger aligns visually with the target prompt, its posture is heavily altered compared to the original.

 \item {Under-editing} (editability $< $ fidelity): This situation arises when the method fails to sufficiently apply the desired changes in the edited regions, maintaining too much of the original image. As a result, the edited image does not accurately reflect the changes specified in the text prompt. For instance, in the last row of \figref{balance}(b), while the structure of the coat is well-preserved, its appearance does not align with the modifications required by the target prompt.
\end{compactitem}
%%%%%%%%%%%%%%%%%%%%%%%%%%%%%%%%%%%%%%%%%%%%%%%%

The existing dual-branch editing paradigm, which mainly utilizes attention injections~\cite{hertz2022prompt,tumanyan2023plug,cao_2023_masactrl,liu2024towards} for structure preservation, \emph{{lacks an explicit method to balance both fidelity and editability.}}
However, adjustments can only be achieved through hyperparameters such as attention injection timesteps.
To address these limitations, we introduce \emph{{UnifyEdit}} to explicitly balance fidelity and editability through \emph{{a unified diffusion latent optimization framework}}, enabling adaptive adjustments to meet the specific requirements of various editing types. 
Specifically, UnifyEdit differs from direct attention injections by employing two attention-based constraints derived from the pre-trained T2I models: the \emph{self-attention (SA) preservation} constraint, which ensures structural fidelity by measuring discrepancies between SA maps of the source and target branches, and the \emph{cross-attention (CA) alignment} constraint, which boosts editability by promoting higher CA values in areas corresponding to the target edited token.

We note that directly combining these two constraints to guide diffusion latent optimization can produce conflicting gradients, causing one constraint to dominate and skew the guidance direction. 
This imbalance may lead to either over- or under-editing failures.
To address this issue, we propose an \emph{adaptive time-step scheduler} that dynamically adjusts the weighting parameters of each constraint according to the denoising timestep.
At the initial denoising stage, when the target diffusion denoising trajectory is close to the source's, emphasis is placed on the CA alignment constraint to enhance editability.
As the denoising process progresses and the target diffusion latent increasingly aligns well with the new prompt, the importance of the SA preservation constraint is heightened to ensure structural fidelity.
Interestingly, we also find that visualizing the gradients of the proposed constraints can pinpoint the causes of over- or under-editing, enabling users to tailor the fidelity-editability trade-off to their preferences.

The main contributions are summarized as follows:
\begin{compactitem}
    \item     
    We introduce UnifyEdit, a novel tuning-free framework that takes the first step toward achieving a balance between fidelity and editability within a \emph{unified} diffusion latent optimization framework.
  
    \item    
We propose an adaptive time-step scheduler that balances two attention-based constraints, one focused on maintaining structural fidelity and the other on enhancing editability. This approach effectively optimizes the diffusion latent toward a balanced direction, accommodating various types of edits.
     \item  
To validate the efficiency of our proposed method in balancing various editing types, we develop a dataset named \emph{Unify-Bench}, which includes a wide range of edits across different scopes of editing regions, such as foreground modifications (changes in color, texture, or material, and object replacements), background editing, global style transfer, and human face attribute editing.
Both quantitative and qualitative experimental results demonstrate that our method significantly improves the trade-off between structure fidelity and editing efficiency compared to existing state-of-the-art approaches.
\end{compactitem}

%The rest of this paper is organized as follows. 
%\secref{related} reviews key works in TIE and the specific technique of diffusion latent optimization.
%
%The preliminary of the dual-branch tuning-free TIE paradigm and the proposed framework are illustrated in \secref{preliminary} and \secref{proposed-method}, respectively. 
%
%\secref{experiments} presents detailed experimental results and analysis. 
%
%Finally, \secref{conclusions} concludes the paper.

%%%%%%%%%%%%%%%%%%%%%%%%%%%%%%%%%%%%%%%%%%%%%%%%

\section{Related Work}
\label{sec:related}
%In this section, we provide a brief overview of key works in TIE, highlight the efforts made by existing methods to balance fidelity and editability, and introduce the specific technique of diffusion latent optimization.
%

\subsection{Text-based Image Editing Using Diffusion Models}
\label{subsec:TIE}
Text-based image editing (TIE) involves modifying input images based on specific text prompts while preserving the integrity of the original content. 
With the emergence of diffusion models~\cite{ho2020denoising,song2020denoising}, numerous approaches have been developed to leverage their capabilities for this task.
Existing TIE methods based on diffusion modes can be broadly divided into three categories: training~\cite{Kim_diffusionclip,MingiAsyrp,brooks2023instructpix2pix,zhang2023magicbrush}, fine-tuning~\cite{ValevskiUniTune,kawar2023imagic,zhang2023sine}, and tuning-free methods~\cite{couairon2022diffedit,hertz2022prompt,tumanyan2023plug,cao_2023_masactrl,qiao2024baret,parmar2023zero,mokady2023null,li2024source,titov2024guide,brack2024ledits}.

\emph{Training-based} methods focus on training a model specifically for a given task using a substantial amount of data to transform a source image into a target image.
Early works~\cite{Kim_diffusionclip,MingiAsyrp} such as CLIPDiffusion~\cite{Kim_diffusionclip} mainly target domain-specific image transformations, for instance, transforming a ``dog'' into a ``bear''.
In particular, it leverages CLIP loss to train the diffusion model that aligns the generated image with the target text.
However, these methods are constrained by their reliance on short phrases to define domains, which limits their ability to fully utilize the flexibility offered by free-form text.
To address this limitation and reduce the need for complex descriptions, recent methods~\cite{brooks2023instructpix2pix, zhang2023magicbrush, guo2023focus, zhang2024hive, ValevskiUniTune } such as InstructPix2Pix~\cite{brooks2023instructpix2pix} introduce editing driven by natural language instructions such as ``add a flower".
InstructPix2Pix~\cite{brooks2023instructpix2pix} employs a fully supervised approach, utilizing training datasets of paired source and edited images with corresponding instructions. 
It enhances the backbone T2I model with an additional input channel for incorporating image conditions during denoising. 
This allows it to produce images that adhere to the instructions while maintaining the original's integrity. 
Furthermore, it adapts Classifier-Free Guidance (CFG)~\cite{dhariwal2021diffusion} to balance alignment with the input image and edit instructions.

To reduce the computational cost associated with training a full diffusion model, \emph{fine-tuning} methods~\cite{kawar2023imagic,ValevskiUniTune,zhang2023sine} focus on refining specific layers or embeddings rather than the entire denoising network.
UniTune~\cite{ValevskiUniTune} fine-tunes the diffusion model on a single base image during the tuning phase, ensuring that the generated images closely resemble the base image.
Imagic~\cite{kawar2023imagic} optimizes the text embedding and fine-tunes the T2I model by minimizing the discrepancy between the reconstructed and original images.
 The final edited image is then generated by linearly interpolating the optimized text embedding with the target text using the fine-tuned diffusion model.

Unlike training or fine-tuning diffusion models, numerous \emph{tuning-free} methods that directly adapt existing pre-trained T2I models for image manipulations have recently gained substantial attention.
The core idea behind these methods is to use the diffusion denoising process of the T2I model to preserve the fidelity of parts of the original image while simultaneously leveraging the inherent editability of the original text-visual alignment.
These approaches can be broadly categorized into two representative methods: inpainting-based methods~\cite{avrahami2022blended,avrahami2023blended,couairon2022diffedit,huang2023pfb,wang2023instructedit} and dual-branch-based methods~\cite{hertz2022prompt,tumanyan2023plug,cao_2023_masactrl,magedit,tang2024locinv,wang2024vision,titov2024guide}.
Inpainting-based methods, such as DiffEdit \cite{couairon2022diffedit}, utilize a mask to merge the noisy image in the edited region, which is guided by text prompts, with the area outside of the mask using the noisy source image.
%
% P2P ~\cite{hertz2022prompt} stands out as one of the pioneering dual-branch-based methods. It introduces the innovative approach of injecting CA and SA maps, derived from the source image, into the target branch.
Recently, the dual-branch P2P~\cite{hertz2022prompt} model extracts self-attention and cross-attention maps from the source image and injects them into the target branch for editing. 
In this work, we focus on \emph{tuning-free} methods, eliminating the need for extensive retraining and thus saving time and computational resources. 

\subsection{Tuning-Free Text-based Image Editing}
In TIE, achieving a balance between fidelity and editability is important to generate high-quality results.
\emph{Fidelity} involves preserving the original content that should not be changed, while \emph{editability} ensures the desired changes that satisfy the text prompt.
As categorized in \subsecref{TIE}, recent tuning-free TIE methods fall into two representative categories: inpainting-based and dual-branch-based methods.
These approaches utilize distinct mechanisms to balance fidelity and editability.
\emph{Inpainting-based} methods~\cite{avrahami2022blended,avrahami2023blended,couairon2022diffedit,huang2023pfb,wang2023instructedit} prioritize preserving fidelity in non-edited regions by introducing advanced mask-grounding and mask-blending techniques.
They aim to accurately identify the target region and seamlessly integrate the generated foreground object with the background latent of the source image, ensuring a natural and cohesive result.
In particular, Blended Latent Diffusion~\cite{avrahami2023blended} directly generates a foreground object based on text prompts and introduces an improved blending operation to seamlessly integrate the object with the background latent of the source image.
DiffEdit~\cite{couairon2022diffedit,wang2023instructedit} proposes an unsupervised mask-predicting method and utilizes DDIM inversion~\cite{song2020denoising} to integrate latent features alongside the target prompt, thereby generating the foreground image.
However, these methods often result in significant structural alterations in the target foreground objects due to the inadequate structural information provided by the source image.

To maintain the overall fidelity of edited images, dual-branch-based 
methods~\cite{hertz2022prompt,tumanyan2023plug,cao_2023_masactrl} such as P2P~\cite{hertz2022prompt} leverage self-attention and cross-attention attention injection from the source branch to guide the target branch. 
%
%Further subdivisions can be made in the dual-branch approach.
%
\emph{Attention-injection-based} methods~\cite{hertz2022prompt,tumanyan2023plug,cao_2023_masactrl,qiao2024baret} emphasize extracting and injecting highly expressive features into the target branch, thereby enhancing structure preservation of the edited images.
Recent advancements in \emph{inversion-based} methods~\cite{mokady2023null,ju2023direct,li2024source} refine the inversion process to enable a more precise source branch of the original image, thereby generating enhanced feature injection sets compared to DDIM inversion~\cite{song2020denoising}.
Motivated by inpainting-based methods, recent dual-branch approaches~\cite{tang2024locinv,wang2024vision,magedit} further utilize masks to focus on preserving fidelity in the non-edited regions and effectively prevent unintended attribute leakage.

Despite their success, existing dual-branch-based methods mainly regulate balance by adjusting attention injection timestep hyperparameters.
However, how to balance both fidelity and editability within a unified framework has been overlooked in the literature.
In this work, we concentrate on explicitly balancing fidelity and editability within a unified diffusion latent optimization framework.
The two works most closely related to ours are Guide-and-Rescale (G-R)\cite{titov2024guide} and MAG-Edit \cite{magedit}, which both employ \emph{gradient-based} methods to formulate either fidelity or editability using attention-based constraints explicitly.
G-R~\cite{titov2024guide} proposes to model the structure preservation using the SA constraint and performs gradient optimization using noise guidance.
However, it merely leverages the text's inherent editability within the CFG without explicitly modeling.
On the other hand, MAG-Edit~\cite{magedit} proposes amplifying the CA values within the mask to locally enhance the text alignment, and improve editability through diffusion latent optimization. 
Nevertheless, it still relies on attention injection for structure preservation.
In contrast, our UnifyEdit explicitly integrates two powerful constraints for both fidelity and editability to guide the diffusion latent in a unified manner adaptively. 

\subsection{Diffusion Latent Optimization}
\label{subsec:latent}
Diffusion latent optimization iteratively refines latent variables during denoising by minimizing specified constraints to align the latent trajectory with the target distribution and guide outcomes toward desired results.
This technique has been effectively used in training-free T2I generation~\cite{chefer2023attend,rassin2024linguistic,xie2023boxdiff,ge2023expressive,dahary2024yourself, LiuComposite}, for improving semantic alignment and enabling training-free control.
Specifically, Attend-and-Excite~\cite{chefer2023attend} and Linguistic Binding~\cite{rassin2024linguistic} utilize latent optimization with cross-attention constraints to address attribute leakage and incorrect binding.
Additionally, latent optimization facilitates training-free condition control, such as color and layout, during the image generation process.
For instance, Rich-Text-to-Image~\cite{ge2023expressive} employs an objective function that minimizes the discrepancy between the predicted initial latent and a predefined RGB triplet, thereby enabling precise control over the color of generated objects.
Similarly, training-free layout generation methods~\cite{xie2023boxdiff,dahary2024yourself, LiuComposite} leverage latent optimization by formulating objectives based on cross-attention maps and bounding boxes, effectively positioning objects within designated regions.

While latent feature optimization has shown its effectiveness in T2I, its application in TIE has received comparatively less attention.
In TIE, Pix2Pix-Zero~\cite{parmar2023zero} leverages latent optimization to minimize discrepancies between the CA maps of the source and target branches, effectively preserving the fidelity of edited images.
Most recently, MAG-Edit~\cite{magedit} utilizes latent optimization to enhance the alignment between textual prompts and latent features, significantly improving editability.
These advancements highlight the potential of diffusion latent optimization in TIE.
Compared to diffusion latent optimization, many editing-based methods focus on utilizing noise guidance~\cite{titov2024guide,diffeditor}.
From the perspective of score-based diffusion~\cite{song2020score}, both noise guidance and latent optimization construct an energy function $g(z_t,y)$ and compute the gradient $\nabla_{z_t}g(z_t,y)$.
However, the key difference lies in their application of the gradient: noise guidance uses it to update the predicted noise $\epsilon_{\theta}$, guiding the sampling of $z_{t-1}$, 
whereas latent optimization directly adjusts $z_t$ itself and recomputes the $\epsilon_{\theta}$ based on the new $z_t$. 
Compared to noise guidance, under this framework, the sampling of $z_t$ can be viewed as a fixed-point problem, solving it iteratively to reach the optimal solution.
In this work, we adopt the diffusion latent optimization technique and present additional comparisons with noise guidance in \subsecref{discussion} using the same constraints.

%%%%%%%%%%%%%%%%%%%%%%%%%%%%%%%%%%%%%%%%%%%%%%%%%%%%%%%%%%%%%%%%%%%%%%%%%%%%%%%%%%
%%%%%%%
%MH: change caption (a) to Experiments on SA constraint and (b) Experiments on CA constraints
\begin{figure*}[!t]
    \centering
    \includegraphics[width=1.0\linewidth]{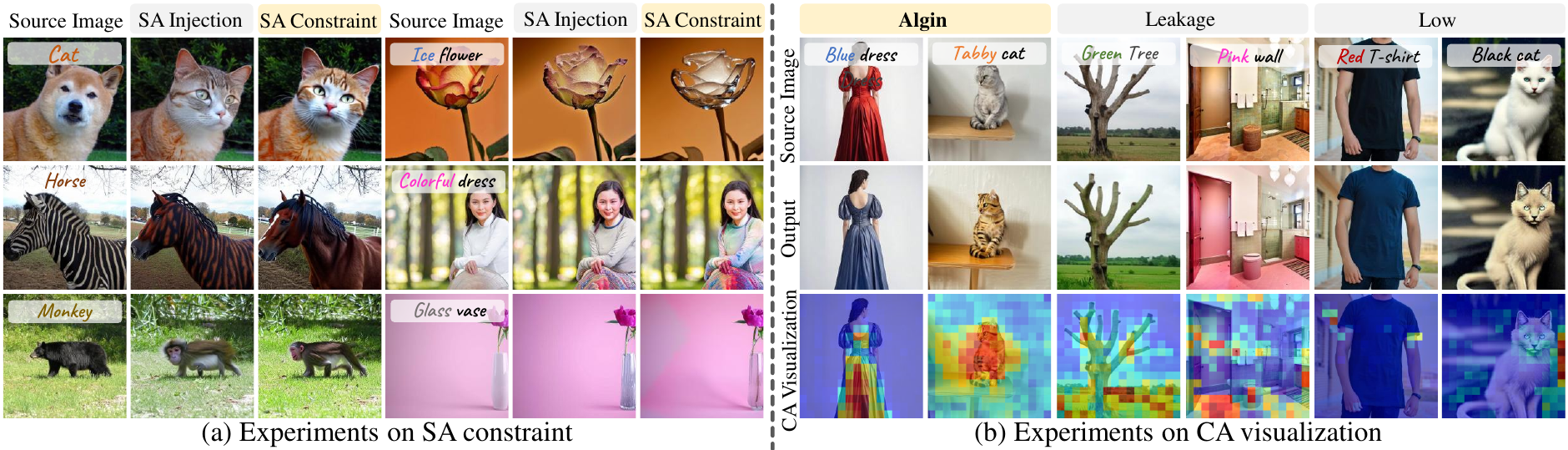} 
    \vspace{-5 mm}
    \caption{
    \textbf{Experiments with self-attention and cross-attention.}
    (a) Compared to SA injection, the SA constraint offers greater flexibility in editing.
    (b) When the CA map accurately focuses on the target region with a strong response, the resulting edits align effectively with the text prompt. However, attention leakage or low attention values can lead to misalignment or ineffective editing outcomes.
    }
    \label{fig:motivation}
    \vspace{-5 mm}
\end{figure*}
%%%%%%%%%%%%%%%%%%%%%%%%%%%%%%%%%%%%%%%%%%%%%%%%

%HERE
\section{Dual-Branch Tuning-Free Image Editing}
\label{sec:preliminary}
The goal of text-based image editing is to transform the source image $\mathcal{I}$ into a target image $\mathcal{I}^\ast$ that aligns with the target prompt $\mathcal{P}^\ast$ while preserving the content of $\mathcal{I}$ that is not intended to be changed.
To achieve this goal, the dual-branch editing paradigm~\cite{hertz2022prompt,tumanyan2023plug,parmar2023zero,mokady2023null,cao_2023_masactrl, magedit,li2024source,titov2024guide} is widely adopted in the literature.
As illustrated in \figref{difference}(a),
this paradigm includes two branches: the source branch, generated by the original prompt $\mathcal{P}$, and the target branch, generated by the target prompt $\mathcal{P}^*$.
The set of new target tokens present in $\mathcal{P^*}$ against $\mathcal{P}$ is defined as $\mathcal{S}^\ast=\{s^*_1, s^*_i,...,s^*_I\}$, for instance, %``{\textcolor[HTML]{C55A11}{\textbf{graffitied}}}''.
``graffitied''.
Both branches begin with the same initial noise latent feature $z_T$ and end with different outputs:
a reconstructed image $\hat{\mathcal{I}}$ in the source branch and an edited image $\mathcal{I}^\ast$ in the target branch.
Existing methods typically focus on improving the following three operations to enhance fidelity and editability.

\Paragraph{Attention Injection.} 
The layer of denoising U-Net in the T2I model, such as Stable Diffusion~\cite{rombach2022high} contains an SA block and a CA block.
The SA block captures long-range interactions between image features, and the CA block integrates visual features with the text prompt.
Both can be uniformly expressed as:
\begin{equation}
\text{Attention}(Q,K,V) =\text{SoftMax}(\frac{QK^\top}{\sqrt d})V, 
\label{attention}  
\end{equation}
where $Q$ is the projected from spatial features, and $K$,$V$ of SA and CA are projected from either the text embedding or spatial features, respectively.
To ensure the overall fidelity of edited images, P2P~\cite{hertz2022prompt} and PnP~\cite{tumanyan2023plug} first propose attention injection, which involves copying the SA maps $A^{self}$ and CA maps $A^{cross}$ generated in the source branch to the target branch. 
Formally, the SA maps $A^{*self}$ and the CA maps $A^{*cross}$ in the target branch can be formulated as:
\begin{equation}
    \begin{split}
    A^{*self} &\gets A^{self}, \\
    A^{*cross} &\gets A^{cross}. 
    \end{split}
\label{eq:attention_injection}  
\end{equation}
With this formulation, numerous methods~\cite{tumanyan2023plug,cao_2023_masactrl,qiao2024baret} have been proposed to identify the most semantically rich features for injection.
However, directly copying these features imposes overly strict conditions, limiting these approaches to achieving balance solely by modulating the attention injection timesteps.

\Paragraph{Advanced Inversion.} To obtain the initial noise latent feature $z_T$, the DDIM inversion scheme~\cite{song2020denoising} is widely adopted, defined as:
 \begin{equation}
 \footnotesize
     \bar{z}_{t+1} = \sqrt{\frac{\alpha _{t+1}}{\alpha_t}}\bar{z}_t +\sqrt{\alpha_{t+1}}(\sqrt{\frac{1}{\alpha_{t+1}}-1}-\sqrt{\frac{1}{\alpha_t}-1})\epsilon_\theta(\bar{z}_t,t,\mathcal{P}),
\label{eq:inversion}  
\end{equation}    
where t = 0, \dots, T-1. 
Using DDIM sampling, the reconstructed latent feature $z_0$ is obtained through the following process:
 \begin{equation}
 \footnotesize
     z_{t-1} = \sqrt{\frac{\alpha _{t-1}}{\alpha_t}}z_t +\sqrt{\alpha_{t-1}}(\sqrt{\frac{1}{\alpha_{t-1}}-1}-\sqrt{\frac{1}{\alpha_t}-1})\epsilon_\theta(z_t,t,\mathcal{P}),
\label{eq:reconstruction}  
\end{equation}     
where $t = T$, \dots, $1$ and $z_T=\bar{z}_T$.
However, the CFG~\cite{ho2022classifier} in T2I models amplifies accumulated errors from DDIM inversion,  resulting in a significant discrepancy between $\bar{z}_t$ and $z_t$.
This deviation shifts the denoising sampling trajectory of the source branch away from the inversion path and propagates to the target branch through attention injection, thereby adversely impacting fidelity.
To address this issue, several approaches~\cite{mokady2023null,ju2023direct,li2024source} either mathematically enhance the DDIM inversion~\cite{ju2023direct,li2024source} or introduce additional supervisory controls~\cite{mokady2023null, cho2023noise} during sampling for more accurate shared features and greater flexibility in achieving editability.

\Paragraph{Editing Area Grounding and Blending.} 
For localized editing, to preserve the fidelity outside the foreground region, the mask grounding and blending operation is typically employed~\cite{avrahami2022blended,hertz2022prompt} to integrate the edited object with the background.
This operation combines the latent noise feature $z_t^\ast$ from the target branch within the grounded editing area $\mathcal{M}$ with $z_t$ from the source branch outside $\mathcal{M}$, given by:
\begin{equation}
   z^\ast_t = \mathcal{M} \odot z^\ast_t + (1-\mathcal{M})\odot z_t
\label{eq:blend}  
\end{equation}    
To seamlessly integrate the content inside and outside the targeted foreground region, automatic mask-grounding methods and better blending operations have been significantly proposed~\cite{avrahami2023blended,couairon2022diffedit,patashnik2023localizing,huang2023pfb,wang2023instructedit,lu2023tf,brack2024ledits,tang2024locinv,wang2024vision}.

\section{Unfiy-Edit via Latent Optimization}
\label{sec:proposed-method}

Although the dual-branch editing paradigm uses attention injection to preserve fidelity, it lacks a systematic method for balancing fidelity and editability.
Existing schemes are primarily limited to tuning hyperparameters~\cite{hertz2022prompt,tumanyan2023plug,cao_2023_masactrl,li2024source}, such as attention injection timesteps~\cite{hertz2022prompt,tumanyan2023plug,li2024source}.
In this work, we explicitly balance fidelity and editability through a unified framework that allows for adaptable modifications tailored to the diverse needs of different editing scenarios.
In contrast to attention injections of the dual-branch paradigm shown in \figref{difference}(a),
we propose \textbf{\emph{UnifyEdit}} that optimizes the diffusion latent $z_t^\ast$ in the target branch guided by two attention-based constraints to achieve a balance between fidelity and editability, as illustrated in \figref{difference}(b). 
Furthermore, we introduce a mask $\mathcal{M}$ for localized editing to target and balance the edited region specifically. 
The mask-guided optimization is formalized as follows:
\begin{equation}
  \hat{z}_t^*= z_t^* - \mathcal{M} \odot\nabla_{z_t^*}\mathcal{L},
\label{eq:proposed5}  
\end{equation}
%
%where $\mathcal{M}$ localizes the target edited region.
%
For global editing,
$\mathcal{M}$ is set to a matrix of all ones.
This formulation enables precise adjustments within specified areas or across the entire image, depending on the requirements.

To formulate $\mathcal{L}$ that guides $z_t^*$ from both fidelity and editability perspectives, we begin by rethinking the roles of SA and CA in TIE, conducting two experiments described in~\subsecref{motivation}.
Specifically, we propose two attention-based constraints to model fidelity and editability, detailed in \subsecref{self_constraints}.
Finally, leveraging the adaptive time-step scheduler introduced in \subsecref{atb}, our framework dynamically balances these constraints to meet specific requirements of various editing tasks.

%%%%%%%%%%%%%%%%%%%%%%%%%%%%%%%%%%%%%%%%%%%%%%%%%%%%%%%%
\begin{figure*}[!t]
    \centering
    \includegraphics[width=0.99\linewidth]{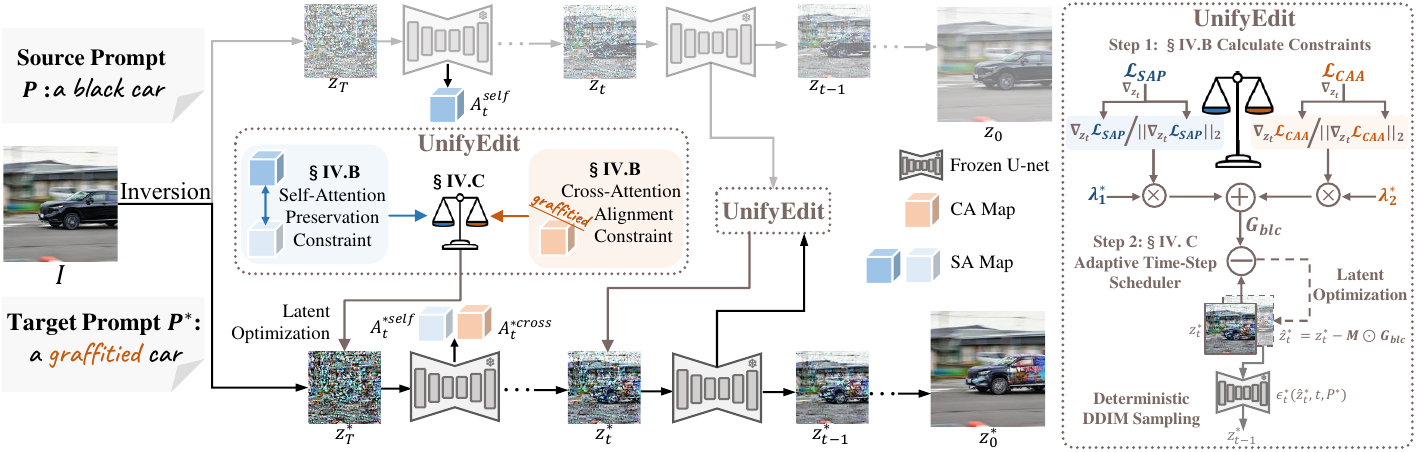}    
    \vspace{-2 mm}
    \caption{\textbf{Illustration of UnifyEdit.}
    UnifyEdit is applied to the diffusion latent feature $z_t^\ast$ in the target branch, involving two key steps: 1) calculating $\mathcal{L}_{\rm{SAP}}$ and $\mathcal{L}_{\rm{CAA}}$ for fidelity and editability, and 2) applying an adaptive time-step scheduler for latent optimization.
    }
    \vspace{-4 mm}
    \label{fig:framework}
\end{figure*}
%%%%%%%%%%%%%%%%%%%%%%%%%%%%

%HERE
\subsection{Rethinking Self- and Cross-Attention for TIE}
%Text-Based Image Editing
\label{subsec:motivation}
For fidelity, previous works~\cite{tumanyan2023plug,liu2024towards} have demonstrated that SA maps play a more significant role in preserving the layout and structure of images than CA maps. 
Regarding editability, existing studies~\cite{guo2023focus,magedit} have demonstrated that the CA maps are crucial for aligning the editing effects with the text prompt.
In this section, we conduct two experiments within the commonly used dual-branch editing paradigm, P2P~\cite{hertz2022prompt}, to better understand how SA and CA influence the editing process in TIE.
Note that all experiments are conducted without CA injection.

\Paragraph{Experiments on Self-Attention:} 
We replace SA injection in P2P~\cite{hertz2022prompt} by optimizing the diffusion latent feature $z_t^\ast$ to minimize the $L_2$ loss between SA maps from the source and target branches.
%

%\Paragraph{Note 1:} 
\emph{Optimizing $z_t^\ast$ with the SA constraint effectively preserves the layout and structural fidelity of the original image while unleashing greater editing flexibility compared to direct SA injection.}
Both the SA injection and the SA constraint effectively preserve the structure and layout fidelity of the original image without requiring CA injection.
However, using the SA constraint allows for greater flexibility in appearance editing. 
For example, it results in the texture of the original ``zebra" fading completely, while the shirt successfully transitions to a  ``colorful" appearance, as illustrated in \figref{motivation}(a).

\Paragraph{Experiments on Cross-Attention:} 
We visualize the average of all CA maps corresponding to the target token (\eg ``\textcolor{blue}{Blue}'' in the first column of ~\figref{motivation}(b)) at a resolution of $16\times16$ for both successful and failed editing examples in P2P~\cite{hertz2022prompt}.

%\Paragraph{Note 2:} 
\emph{High-response CA values indicate strong alignment between text and image features, resulting in pronounced editing effects.}
As demonstrated in the first two columns of \figref{motivation}(b), when the CA map accurately focuses on the intended region, the editing output aligns effectively with the textual prompt. 
Conversely, misaligned or weak values of CA maps lead to editing leakage or under-editing issues. 
For instance,
in the ``green tree'' scenario, excessive attention leakage to the ground areas in the CA map of token ``green'' results in edits mistakenly targeting the ground rather than the tree itself. 
Additionally, the low CA responses for the ``red'' token in the T-shirt area result in minimal changes to the T-shirt’s color to red.
Thus, CA maps inherently reflect the degree of text-visual alignment, and controlling them offers a promising approach to enhancing editability.

%MH: do not need to use the word toy (does not sound good)
%HERE
\subsection{Deriving Attention-Based Constraints}
\label{subsec:self_constraints}
Based on the experimental results discussed above, we introduce two constraints leveraging SA and CA to explicitly model fidelity and editability, respectively, as shown in \figref{framework}.

\Paragraph{Self-Attention Preservation Constraint.}
As demonstrated in \subsecref{motivation}, using the constraint to reduce the discrepancy between the SA maps $A^{self}_t$ generated from the source branch and $A^{*self}_t$ from the target branch successfully preserves the structural fidelity and offers more flexibility than direct SA injection.
The SA preservation constraint is defined as:
\begin{equation}
\mathcal{L}_{\rm{SAP}}=\sum(A^{self}_t-A^{*self}_t)^2.
\label{proposed1}  
\end{equation}
Furthermore, editing small objects within complex scenarios requires more precise control to preserve high-frequency details.
In such cases, a region-based SA preservation constraint proves more effective: 
\begin{equation}
\mathcal{L}_{\rm{R-SAP}}=\sum(\hat{\mathcal{M}} \odot A^{self}_t-\hat{\mathcal{M}} \odot A^{*self}_t)^2,
\label{proposed2}  
\end{equation}
where the mask $\hat{\mathcal{M}}$ for the SA maps is defined as
$\hat{\mathcal{M}}= \boldsymbol{\mathcal{M}}\boldsymbol{\mathcal{M}}^\top$,
with $\boldsymbol{\mathcal{M}}$ representing the flattened vector of $\mathcal{M}$.
Although full-resolution SA maps are generally used to construct this constraint, for tasks requiring significant shape variation (\eg object replacement), we specifically employ SA maps at $16 \times 16$ and $8 \times 8$ resolutions.
Notably, our method is compatible with various inversion techniques.
Furthermore, the SA maps can be directly derived from DDIM inversion~\cite{song2020denoising} rather than from a dedicated denoising sampling process in the source branch.
We discuss them in \subsecref{discussion} and present the experimental results in~\figref{fatezero}.

\Paragraph{Cross-Attention Alignment Constraint.}
\label{constraint}
Larger and more aligned CA values signify stronger alignment of text-visual features, thereby enhancing editability in intended regions, as demonstrated in \subsecref{motivation}. 
Accordingly, we design the CA alignment constraint to maximize the ratio of CA values for the targeted token within the predefined editing region $\mathcal{M}$ relative to those outside $\mathcal{M}$.
Consider the CA map $(A^{*cross}_t)_i$ corresponding to the $i$-th editing token $\mathcal{S}^*_i$ (\eg ``graffitied'' 
in~\figref{framework}(a)) within a mask region $\mathcal{M}$.
The constraint emphasizes increasing the proportion $R_l$ of the $\mathcal{S}^*_i$’s CA values within $\mathcal{M}$ relative to those outside mask $1-\mathcal{M}$:
\begin{equation}
R_l = \frac{\frac{1}{|\mathcal{M}|} \sum\limits_{j \in \mathcal{M}} (A^{*cross}_t)_i^l}{\frac{1}{|1-\mathcal{M}|} \sum\limits_{j \notin \mathcal{M}} (A^{*cross}_t)_i^l},
\label{proposed3}  
\end{equation}
where $j$ denotes the spatial index of CA maps, $l$ represents the CA maps at the $l$-th layer, and $|\ast|$ calculates the total number inside or outside of the mask.
All CA maps are computed in
the resolution of $16\times16$ of the U-Net, recognized for containing the most semantically rich information~\cite{chefer2023attend}.
Furthermore, our experiments reveal that promoting the ratio $R_l$ in each layer at this resolution can further encourage alignment.
Consequently, the CA aligned constraint is defined as:
\begin{equation}
\mathcal{L}_{\rm{CAA}} = -\left(\sum_{l=1}^{L} \sqrt{R_l}\right)^2,
\label{proposed4}  
\end{equation}
where $L = 5$, representing the total number of CA layers at a resolution of $16\times16$.
%

%%%%%%%%%%%%%%%
\begin{figure*}[!t]
    \centering
    \includegraphics[width=1.0\linewidth]{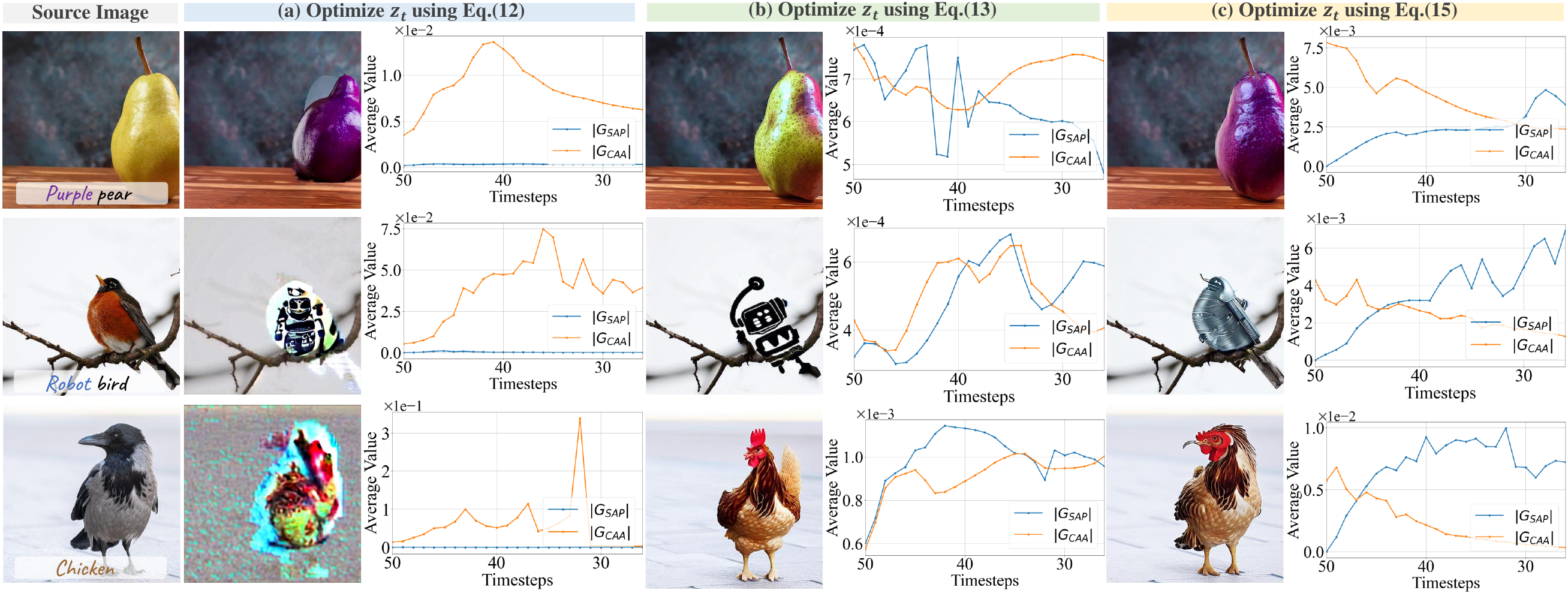}    
    \vspace{-5 mm}
    \caption{\textbf{Editing and visualization results of different gradients.}
    (a) Using \eqnref{a2} alone results in a significantly stronger influence of $\mathcal{L}_{\rm{CAA}}$, disabling $\mathcal{L}_{\rm{SAP}}$ and causing an unbalanced guidance on $z_t$. 
    (b) Although calculating their norms as in \eqnref{a3} brings the magnitudes of the constraints closer, the irregular dynamics lead to either under-editing or over-editing failures.
    (c) In contrast, applying the adaptive time-step scheduler in \eqnref{a4} shapes the gradient trends in \eqnref{a5} such that $\nabla_{z_t^*}\mathcal{L}_{\rm{SAP}}$ starts small and gradually increases, whereas $\nabla_{z_t^*}\mathcal{L}_{\rm{CAA}}$ exhibits the opposite trend, facilitating fidelity-editability balance.
    }
    \vspace{-4 mm}
    \label{fig:atb_vis}
\end{figure*}
%%%%%%%%%%%%%%%%%%%%%%%%%%%%
%%%%%%%%%%%%%%%%%%%%%%%%%%%%%%%%%%%%%%%%%
\setlength{\textfloatsep}{5pt}
\begin{algorithm}[!t]
\caption{A Denoising Step Using UnifyEdit}
\label{update}
\SetAlgoLined
\KwIn{An original and edited prompt $\mathcal{P}$, $\mathcal{P}^*$; a timestep $t$ and corresponding noise latent features of source and target branches $z_t$, $z_t^*$; a maximum iteration step MAX\_IT and the hyperparameters $\beta_1$,$\beta_2$, $k_1$, $k_2$;
a function $\mb{\mathcal{F}_{1}}(\cdot)$ and a a function $\mb{\mathcal{F}_{2}}(\cdot)$ for computing the proposed constraint $\mathcal{L}_{\rm{SAP}}$ and $\mathcal{L}_{\rm{CAA}}$; a pre-trained Stable Diffusion model $SD$.}
\KwOut {the noisy latent feature $z^*_{t-1}$ for the next timestep of the target branch.}
$\lambda_1 = \beta_1 e^{-k_1t}$  \;
$\lambda_2 = \beta_2(1- e^{-k_2t})$ \;
\For{$i=1$ \KwTo $\rm{MAX\_IT}$}
{
$\_, A_t^{self},\_ \gets SD(z_t,\mathcal{P},t)$ \;
$\_, A_t^{\ast self},A_t^{\ast cross} \gets SD(z^*_t,\mathcal{P}^*,t)$ \;
%$\hat{A}_t \gets Inject(A_t,A^*_t)$\tcp*[r]{Operation in P2P.}
    \hspace{0em}\begin{minipage}[b]{0.35\textwidth}
        \begin{tcolorbox}[width=1.0\textwidth,height=2.3cm,valign=center,colback=white, boxrule=0.5pt,boxsep=0.0mm,left=0.5 mm,right=0.1mm,sharp corners,]
          $\mathcal{L}_{\rm{SAP}} \gets \mathcal{F}_{1}(A_t^{self},A_t^{\ast self}) $\;
          $\mathcal{L}_{\rm{CAA}} \gets \mathcal{F}_{2}(A_t^{\ast cross}) $\;
          $\mathcal{G}_{blc} = \lambda_1^\ast\frac{\nabla_{z_t^*}\mathcal{L}_{\rm{SAP}}}{||\nabla_{z_t^*}\mathcal{L}_{\rm{SAP}}||_2} + \lambda_2^\ast\frac{\nabla_{z_t^*}\mathcal{L}_{\rm{CAA}}}{||\nabla_{z_t^*}\mathcal{L}_{\rm{CAA}}||_2}$ \;
          $\hat{z}_t^\ast=z_t^\ast - \mathcal{M}\odot\mathcal{G}_{blc}$\;
        \end{tcolorbox}
    \end{minipage}
    \hfill
    \hspace{1 mm}\begin{minipage}[t]{0.25\textwidth}
       \vspace{-4 mm}
         \tcp*[f]{\textcolor{blue}{{UnifyEdit.}}}
    \end{minipage}
}

$z^*_{t-1} \gets SD(\hat{z}^*_t, \mathcal{P}^*,t)$ \;
\textbf{Return} $z^*_{t-1}$
\end{algorithm}

\subsection{Balancing via Adaptive Time-Step Scheduler}
\label{subsec:atb}
After obtaining $\mathcal{L}_{\rm{SAP}}$ and $\mathcal{L}_{\rm{CAA}}$ in \subsecref{self_constraints},
the simplest formulation of $\mathcal{L}$ in~\eqnref{proposed5} is:
\begin{equation}
\mathcal{L} = \lambda_1\mathcal{L}_{\rm{SAP}} + \lambda_2\mathcal{L}_{\rm{CAA}},
\label{eq:a1}
\end{equation}
where $\lambda_1$ and $\lambda_2$ are static balancing weights.
However, this naive combination of constraints frequently produces unsatisfied editing results and causes image collapse when using consistent $\lambda_*$ values.

We visualize the gradient of $\mathcal{L}$ as follows and analyze the trends of the two gradients:
\begin{equation}
\mathcal{G}_{naive} = \lambda_1 \nabla_{z_t^*}\mathcal{L}_{\rm{SAP}} + \lambda_2 \nabla_{z_t^*}\mathcal{L}_{\rm{CAA}}. 
\label{eq:a2}
\end{equation}
As illustrated in~\figref{atb_vis}(a), since the $\nabla\mathcal{L}_{\rm{CAA}}$ is substantially larger than $\nabla\mathcal{L}_{\rm{SAP}}$, the impact of the latter is significantly diminished, leading to over-editing or image collapse issues.
To mitigate this imbalance, we first propose normalizing the two gradients using their $L_2$ norm as:
\begin{equation}
\mathcal{G}_{norm} =\lambda_1\frac{\nabla_{z_t^*}\mathcal{L}_{\rm{SAP}}}{||\nabla_{z_t^*}\mathcal{L}_{\rm{SAP}}||_2} +  \lambda_2\frac{\nabla_{z_t^*}\mathcal{L}_{\rm{CAA}}}{||\nabla_{z_t^*}\mathcal{L}_{\rm{CAA}}||_2}.
\label{eq:a3}
\end{equation}

Although~\eqnref{a3} brings the effects of both constraints to a similar magnitude, the dynamics of the normalized gradients remain irregular, resulting in unstable editing outcomes. 
Consequently, both under-editing and over-editing occur, as illustrated in~\figref{atb_vis}(b).

Furthermore, we manually adjust $\lambda_1$ and $\lambda_2$ and observe that in successful editing cases, $\nabla_{z_t^*}\mathcal{L}_{\rm{SAP}}$ initially starts small and gradually increases throughout the denoising process, whereas $\nabla_{z_t^*}\mathcal{L}_{\rm{CAA}}$ exhibits the opposite trend. 
These results can be explained as follows. 
During the early stages of denoising, the target and source diffusion trajectories remain relatively close, as they originate from the same last latent feature $z_T$.
This proximity requires a small $\nabla_{z_t^*}\mathcal{L}_{\rm{SAP}}$ and a large $\nabla_{z_t^*}\mathcal{L}_{\rm{CAA}}$ to enhance editability. 
As the diffusion denoising stage moves forward, the latent feature of the target branch progressively aligns with the new target prompt, necessitating an increase in $\nabla_{z_t^*}\mathcal{L}_{\rm{SAP}}$ to preserve structural fidelity.

To enforce this desired gradient behavior, we propose an \emph{Adaptive Time-Step Scheduler} which replaces the constants $\lambda_1$ and $\lambda_2$ with dynamic values $\lambda_1^\ast$ and $\lambda_2^\ast$:
\begin{equation}
\left\{
     \begin{array}{lr}
     \lambda_1^\ast = \beta_1(1- e^{-k_1(T-t)}),&  \\
     \lambda_2^\ast = \beta_2 e^{-k_2(T-t)},&
     \end{array}
\right.
\label{eq:a4}
\end{equation}
where the scaling factors $\beta_1$ and $\beta_2$ control the baseline values of the gradients, influencing the magnitude of $\mathcal{L}_{\rm{SAP}}$ at the endpoint and $\mathcal{L}_{\rm{CAA}}$ at the starting point of the optimization process.
The rate factors $k_1$ and $k_2$ determine the rates at which the gradients rise and decay, respectively.
The variable $t \in \{T, \dots, 1\}$,  indicates the timestep of the denoising sampling process in the T2I diffusion model, allowing the weighting of each constraint to be dynamically adjusted based on the timestep $t$.
We define the adaptive time-step gradient $\mathcal{G}_{blc}$ as:
%%%%%%%%%%%
\begin{equation}
\mathcal{G}_{blc} =\lambda_1^\ast\frac{\nabla_{z_t^*}\mathcal{L}_{\rm{SAP}}}{||\nabla_{z_t^*}\mathcal{L}_{\rm{SAP}}||_2} +  \lambda_2^\ast\frac{\nabla_{z_t^*}\mathcal{L}_{\rm{CAA}}}{||\nabla_{z_t^*}\mathcal{L}_{\rm{CAA}}||_2}.
\label{eq:a5}
\end{equation}
Consequently, \eqnref{proposed5} can be re-written as follows:
\begin{equation}
  \hat{z}_t^*= z_t^* - \mathcal{M} \odot\mathcal{G}_{blc}.
\label{eq:a6}  
\end{equation}
\figref{atb}(c) demonstrates that applying the adaptive time-step scheduler, implemented with \eqnref{a5}, effectively shapes gradient trends as intended,
thereby achieving a better balance between fidelity and editability.
The main steps of UniyEdit are summarized in~\algref{update}.
This tuning-free, inference-stage optimization method is adaptable and can seamlessly apply to any pre-trained T2I models.
It is important to note that parameters $\beta_1$, $\beta_2$, $k_1$, and $k_2$ can be adjusted to balance fidelity and editability, accommodating different editing requirements and tailoring to users' preferences.

\begin{table*}[t]
\begin{center}
\renewcommand{\arraystretch}{1.5}  % Increase row height
\begin{tabular}{ c  c  p{2.0 cm}<{\centering} c c p{1.5cm}<{\centering} }
\toprule
\multicolumn{2}{c}{\textbf{Edit Type}} & \textbf{Source Image} & \textbf{Source Prompt} & \textbf{Target Prompt} & \textbf{Mask} \\
\midrule
 &  & \begin{minipage}[c]{0.11\columnwidth}
    \centering
    \raisebox{-.5\height}{\includegraphics[width=0.92\linewidth]{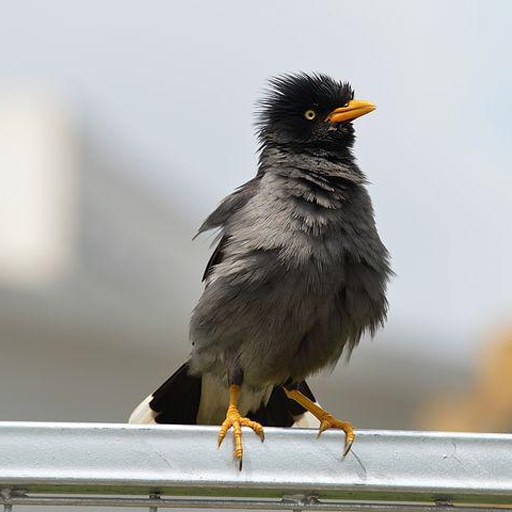}} 
    \end{minipage} 
    & a black bird & a white bird 
    & \begin{minipage}[c]{0.11\columnwidth}
    \centering
    \raisebox{-.5\height}{\includegraphics[width=0.92\linewidth]{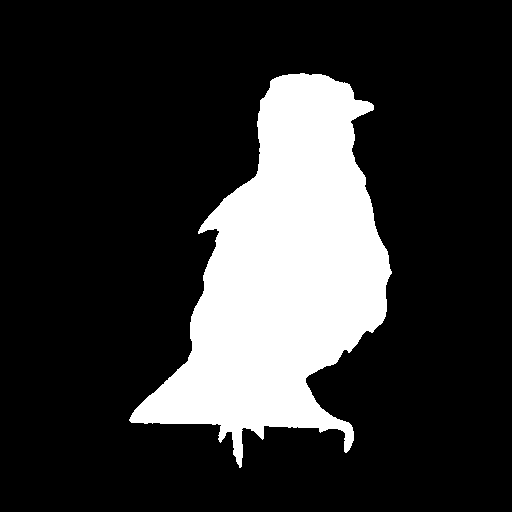}} 
    \end{minipage}    \\
 & \multirow{-3}{*}{\begin{tabular}[c]{@{}c@{}}Color\\ Change\end{tabular}} 
 & \cellcolor[HTML]{EFEFEF} 
    \begin{minipage}[c]{0.11\columnwidth}
    \centering
    \raisebox{-.5\height}{\includegraphics[width=0.92\linewidth]{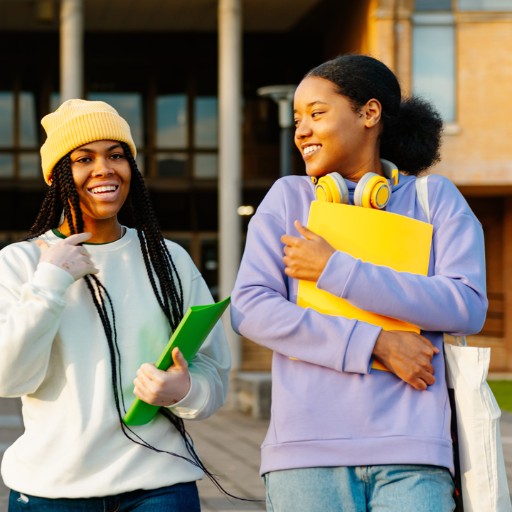}} 
    \end{minipage} 
    & \cellcolor[HTML]{EFEFEF}\begin{tabular}[c]{@{}c@{}}there is a girl wearing\\ a purple beanie\end{tabular} 
    & \cellcolor[HTML]{EFEFEF}\begin{tabular}[c]{@{}c@{}}there is a girl wearing\\ a yellow beanie\end{tabular} 
    & \cellcolor[HTML]{EFEFEF}
    \begin{minipage}[c]{0.11\columnwidth}
    \centering
    \raisebox{-.5\height}{\includegraphics[width=0.92\linewidth]{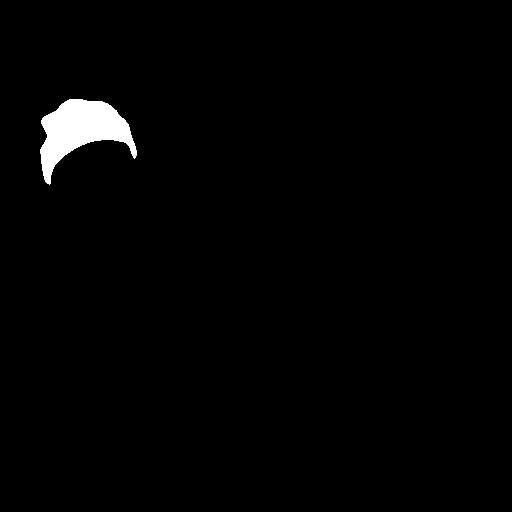}} 
    \end{minipage} \\
 &  & \begin{minipage}[c]{0.11\columnwidth}
    \centering
    \raisebox{-.5\height}{\includegraphics[width=0.92\linewidth]{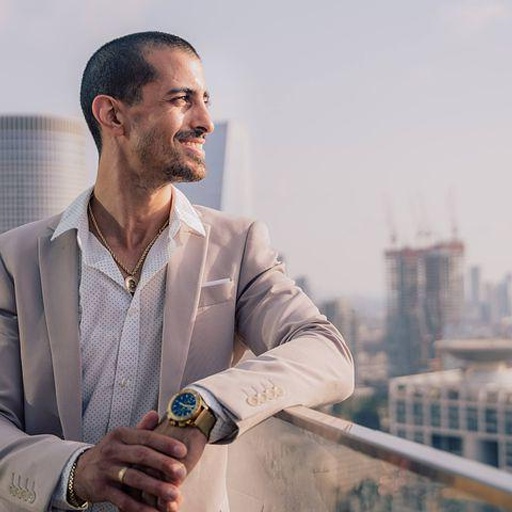}} 
    \end{minipage} 
    & a man in a beige suit & a man in a lace suit 
    & \begin{minipage}[c]{0.11\columnwidth}
    \centering
    \raisebox{-.5\height}{\includegraphics[width=0.92\linewidth]{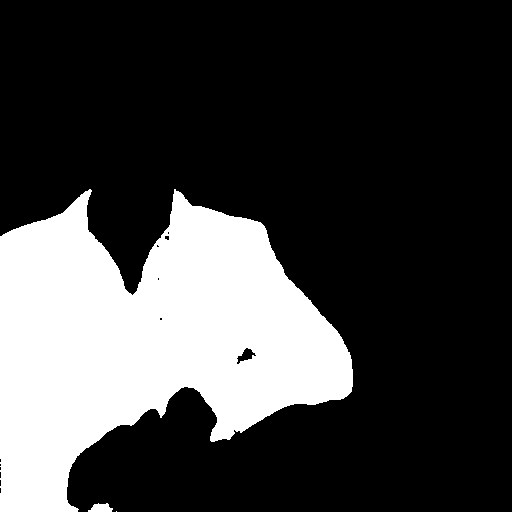}} 
    \end{minipage}   \\
 & \multirow{-3}{*}{\begin{tabular}[c]{@{}c@{}}  Texture \\ Modification\end{tabular}} 
 & \cellcolor[HTML]{EFEFEF} 
    \begin{minipage}[c]{0.11\columnwidth}
    \centering
    \raisebox{-.5\height}{\includegraphics[width=0.92\linewidth]{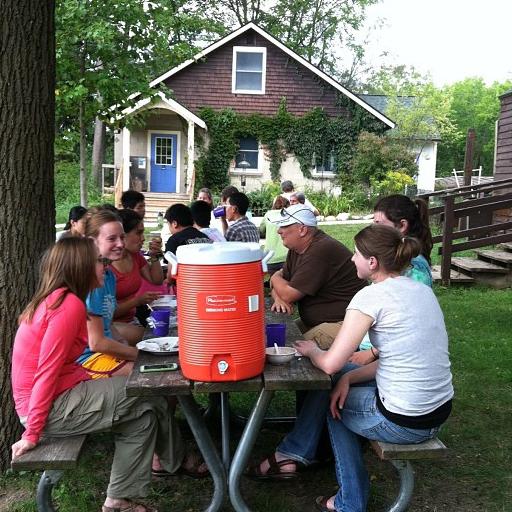}} 
    \end{minipage} 
    & \cellcolor[HTML]{EFEFEF}\begin{tabular}[c]{@{}c@{}}there is a woman in green\\ pants sitting at a table\end{tabular} 
    & \cellcolor[HTML]{EFEFEF}\begin{tabular}[c]{@{}c@{}}there is a woman in wool\\ pants sitting at a table\end{tabular} 
    & \cellcolor[HTML]{EFEFEF}
    \begin{minipage}[c]{0.11\columnwidth}
    \centering
    \raisebox{-.5\height}{\includegraphics[width=0.92\linewidth]{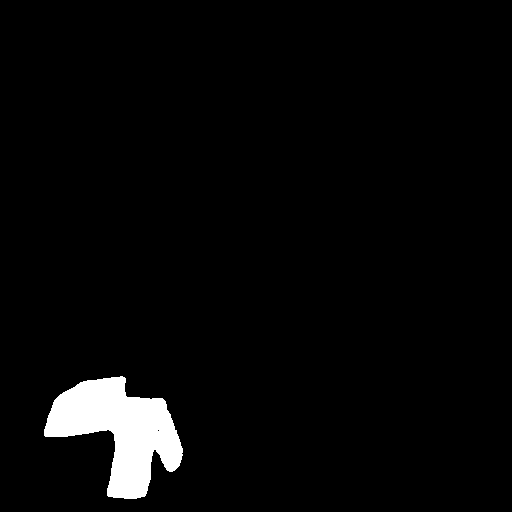}} 
    \end{minipage} \\
  \multirow{-6}{*}{\textbf{\begin{tabular}[c]{@{}c@{}}Foreground\\ \\Editing\end{tabular}}} 
 &  
 & \begin{minipage}[c]{0.11\columnwidth}
    \centering
    \raisebox{-.5\height}{\includegraphics[width=0.92\linewidth]{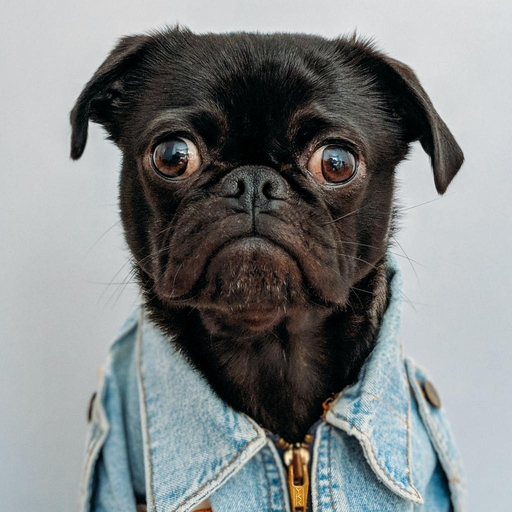}} 
    \end{minipage}   
 & a dog & a fox 
 & \begin{minipage}[c]{0.11\columnwidth}
    \centering
    \raisebox{-.5\height}{\includegraphics[width=0.92\linewidth]{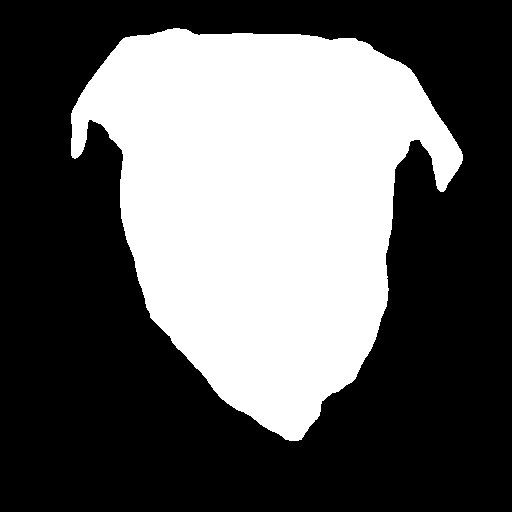}} 
    \end{minipage}    \\
 & \multirow{-3}{*}{\begin{tabular}[c]{@{}c@{}}Object\\ Replacement\end{tabular}} 
 & \cellcolor[HTML]{EFEFEF} 
    \begin{minipage}[c]{0.11\columnwidth}
    \centering
    \raisebox{-.5\height}{\includegraphics[width=0.92\linewidth]{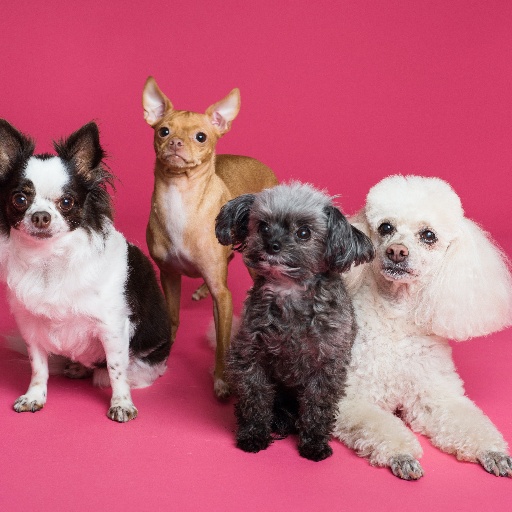}} 
    \end{minipage} 
    & \cellcolor[HTML]{EFEFEF}\begin{tabular}[c]{@{}c@{}}there is a dog standing\\ on a pink background\end{tabular} 
    & \cellcolor[HTML]{EFEFEF}\begin{tabular}[c]{@{}c@{}}there is a cat standing\\ on a pink background\end{tabular} 
    & \cellcolor[HTML]{EFEFEF}
    \begin{minipage}[c]{0.11\columnwidth}
    \centering
    \raisebox{-.5\height}{\includegraphics[width=0.92\linewidth]{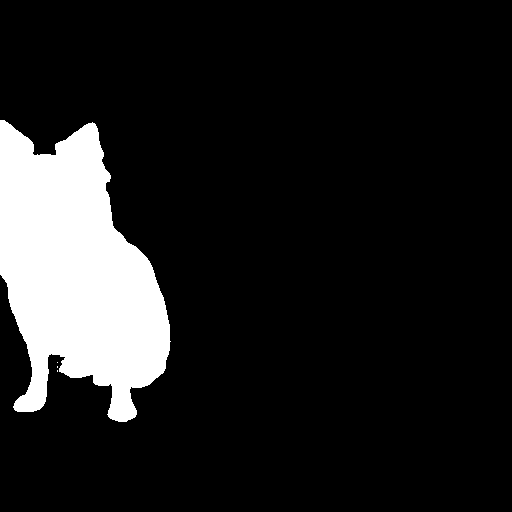}} 
    \end{minipage} \\
\textbf{Background Editing} & \textbackslash
 & \begin{minipage}[c]{0.11\columnwidth}
    \centering
    \raisebox{-.5\height}{\includegraphics[width=0.92\linewidth]{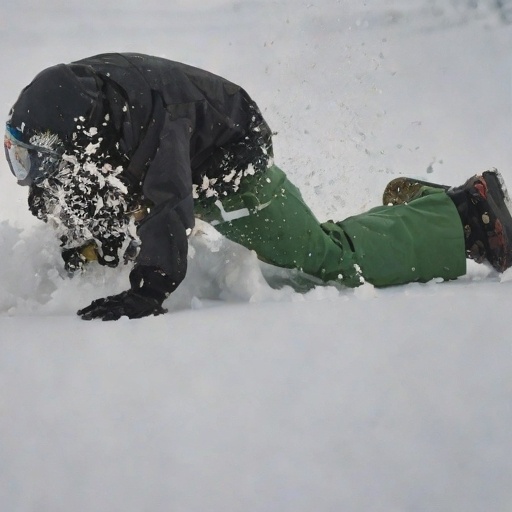}} 
    \end{minipage}  
 & a man falls on snow & a man falls on grass 
 & \begin{minipage}[c]{0.11\columnwidth}
    \centering
    \raisebox{-.5\height}{\includegraphics[width=0.92\linewidth]{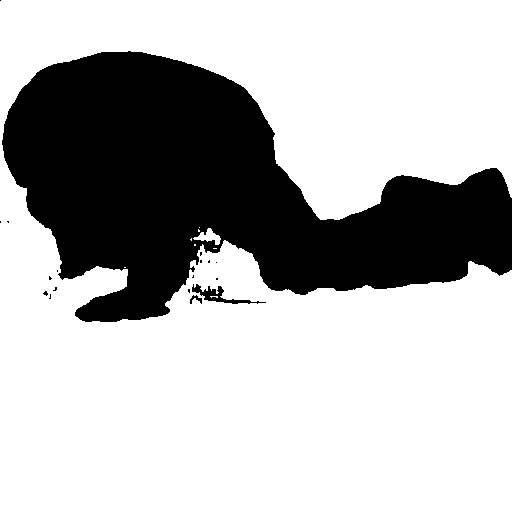}} 
    \end{minipage}   \\[9pt]
\textbf{Global Style Transfer} &\textbackslash
 & 
 \begin{minipage}[c]{0.11\columnwidth}
    \centering
    \raisebox{-.5\height}{\includegraphics[width=0.92\linewidth]{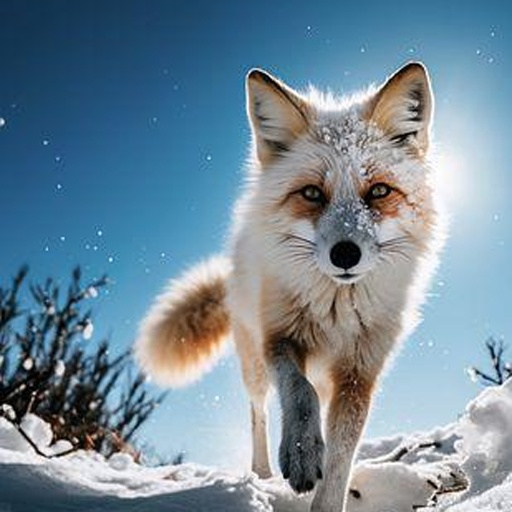}} 
    \end{minipage} 
    & a photo of a fox
    & a drawing of a fox
    &  None  \\[9pt]
\multirow{-1.5}{*}{\begin{tabular}[c]{@{}c@{}}\textbf{Human Face} \\ \textbf{Attribute Editing}\end{tabular}}  &\textbackslash
&\begin{minipage}[c]{0.11\columnwidth}
    \centering
    \raisebox{-.5\height}
    {\includegraphics[width=0.92\linewidth]{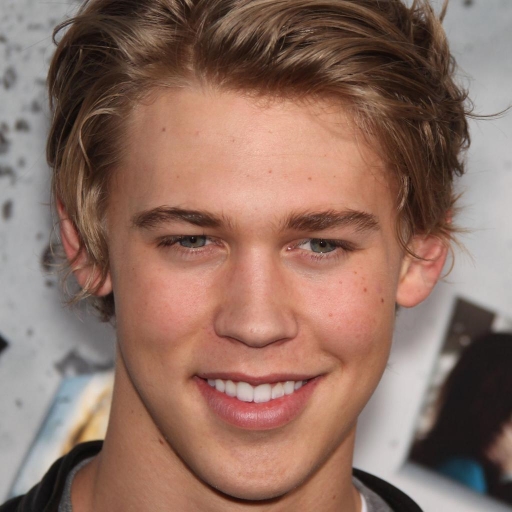}} 
    \end{minipage} 
& a photo of a smiling man & a photo of a crying man
&\begin{minipage}[c]{0.11\columnwidth}
    \centering
    \raisebox{-.5\height}
    {\includegraphics[width=0.92\linewidth]{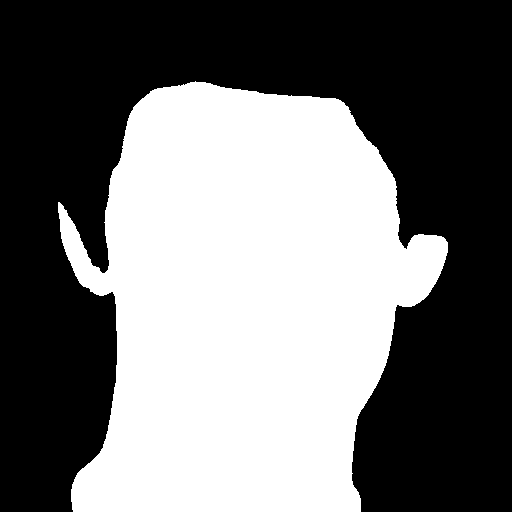}} 
    \end{minipage} \\[5pt]
\bottomrule
\end{tabular}
\end{center}
\vspace{-3 mm}
\caption{\textbf{Examples in Unify-Bench.}
Each image in Unify-Bench is annotated with a source prompt, a target edit prompt, and an edit region mask.
Complex scenarios within the dataset are distinctly highlighted with a  \raisebox{0.3ex}{\colorbox{gray!15}{\strut grey}}.}
\label{tab:dataset}
\vspace{-4 mm}
\end{table*}
%%%%%%%%%%%%%%%%%%%%%%%%%%%%%%%%%%%%
%
%%%%%%%%%%%%%%%%%%%%%%%%%%%%%%%%%%%%%%%%%%%%%%%%%%%%%%%%

\begin{figure*}[!h]
    \centering
    \includegraphics[width=0.98\linewidth]{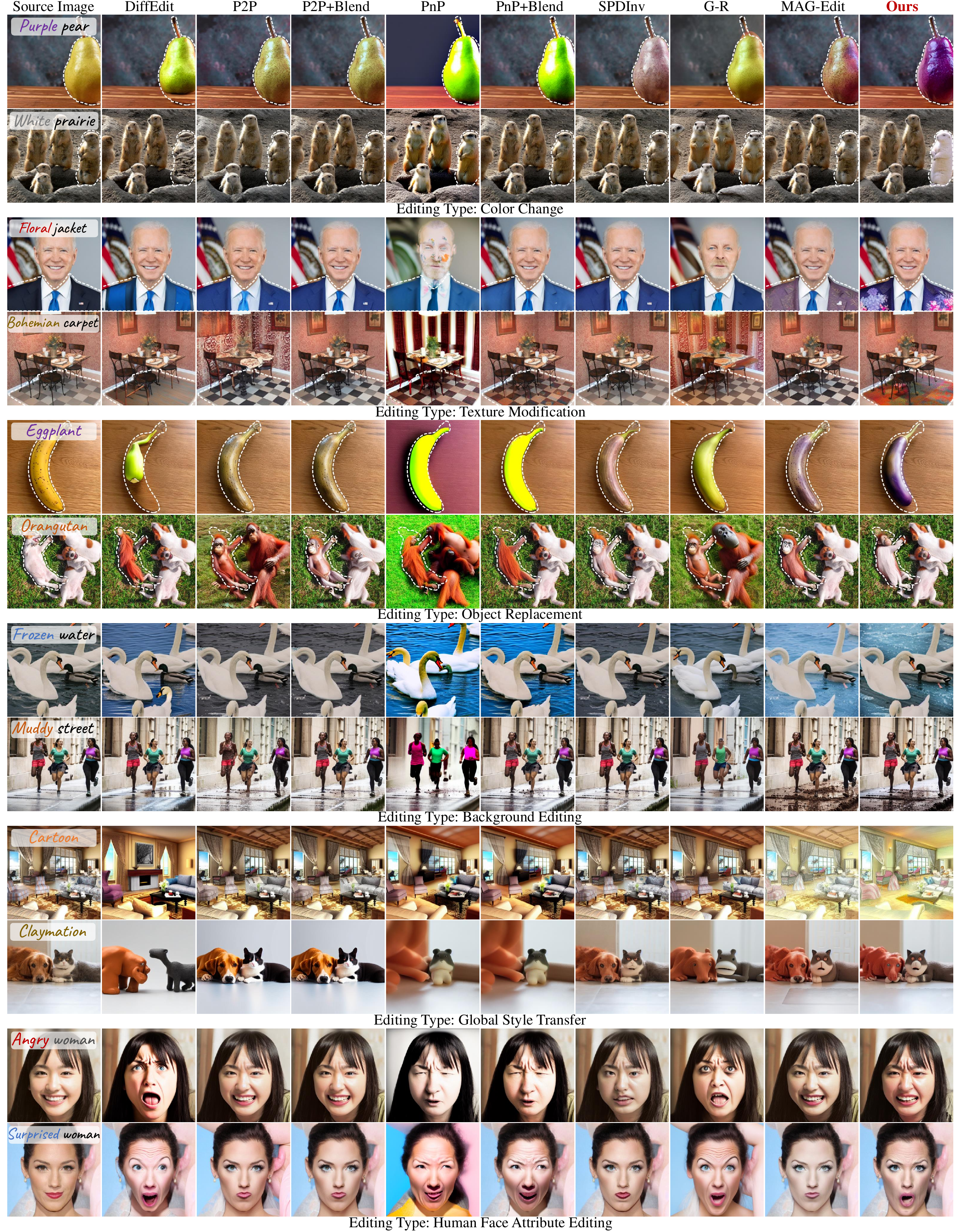}
   \vspace{-2.5 mm}
    \caption{
   \textbf{Qualitative comparisons across various editing types.}
   We use white dashed outlines to highlight the target object in foreground editing.
   Our proposed method achieves a superior balance compared to other baseline methods, demonstrating enhanced editing effects while more effectively maintaining structural consistency. }
    \label{fig:qualitative}
\end{figure*}
%\clearpage
%%%%%%%%%%%%%%%%%%%%%%%%%%%%

%%%%%%%%%%%%%%%%%%%%%%%%%%%%%%%%%%%%
\section{Experiments}
\label{sec:experiments}
\subsection{Benchmark Dataset}
To facilitate comprehensive evaluations of our method's ability to balance fidelity and editability across different editing types, we develop a benchmark dataset named \emph{\textbf{Unify-Bench}}.
This dataset comprises $181$ images sourced from TEd-Bench~\cite{kawar2023imagic}, PIE-Bench~\cite{ju2023direct}, Magicbrush~\cite{zhang2023magicbrush}, and the Internet. 
Unify-Bench is designed to assess the editing capabilities of various methods across different editing regions. 
It includes a diverse range of edits such as foreground modifications, background alterations, global style transfers, and specialized human face attribute editing tasks:
\begin{compactitem}
    \item \textbf{Foreground editing}: it encompasses color change, texture modification, and object replacement. 
    These edits are applied to simple scenarios, which feature a single prominent object, and complex scenarios, which are characterized by multiple objects of the same kind arranged in intricate layouts. In complex scenarios, edits are specifically targeted at a single object.
    
    \item \textbf{Background editing}: it focuses on replacing or modifying the scene behind the foreground subjects.
    
    \item \textbf{Global style transfer}: it aims to globally change the visual style of an image while preserving its underlying content.
    
    \item \textbf{Human face attribute editing}: it includes changing facial expressions, hair color, gender, age, and \etc.
\end{compactitem}
For generating source and target prompts, we initially utilized GPT-4~\cite{OpenAI2023GPT4TR}, followed by manual refinement to ensure the accuracy and relevance of these prompts.
For conciseness, we use the simplest possible prompts, employing the format ``a XX'' for simple scenarios and ``there is XX in/on XX'' for complex scenarios.
For localized editing, we generate the corresponding editing masks using the Segment Anything method~\cite{facebook2:online}. 
Thus, each image in Unify-Bench is annotated with a source prompt, a target edit prompt, and an edit region mask, as detailed in~\tabref{dataset}. 

\subsection{Implementation Details}
\label{subsec:implementation_details}
All experiments are conducted on a single NVIDIA A100 GPU.
We utilize the official pre-trained Stable Diffusion v1.4 model~\cite{CompViss92:online} as our base model.
%\footnote{\url{https://github.com/CompVis/stable-diffusion}} 
%
We choose Null-Text inversion (NTI)~\cite{mokady2023null} as the inversion method to obtain $z_T$, and the denoising sampling process employs the DDIM method~\cite{song2020denoising} over $T=50$ steps, maintaining a constant CFG scale of $7.5$. 
Our approach is compatible with various inversion methods (see~\subsecref{discussion}).
The $\mathcal{L}_{\rm{CAA}}$, $\mathcal{L}_{\rm{SAP}}$ are applied during diffusion steps within the ranges $[T,\tau_1]$ and $[T,\tau_2]$, respectively, with both $\tau_1$ and $\tau_2$ typically set at $25$.
However, $\tau_1$ is specifically set at $5$ for color editing. 
We generally set the scaling factors $\beta_1$ and $\beta_2$ to $5$.
For rate factors, we set $k_1$ and $k_2$ as follows: 
$0.05$ for color change, $0.08$ for texture modification or background editing, $0.15$ for object replacement, $0.1$ for global style transfer and $0.25$ for human face attribute editing.
The optimization process is further defined by the maximum number of iterations, empirically set to $\rm{MAX\_IT}=1$. 
For localized editing, we employ the mask blending operation in P2P~\cite{hertz2022prompt} with the annotated masks to better preserve the original information in areas outside the mask.

\subsection{Evaluation Metrics}
We quantitatively evaluate our proposed
method against baseline models using both automatic metrics and
human evaluations.

\Paragraph{Automatic Metrics.} 
We assess the efficiency of our method in terms of fidelity and editability using the following automatic metrics:
1) \textbf{fidelity}:
We calculate the DINO-ViT self-similarity (DINO Similarity)~\cite{omerbtSp27:online} between the source and edited images to analyze structure preservation.
2) \textbf{editability}:
We compute the CLIP score~\cite{showlabl4:online} with the CLIP ViT-L/14 model by evaluating the similarity between text and image embeddings in a shared space to measure image-text alignment.
We adapt the reference code for localized editing to crop the target regions in both source and edited images using bounding boxes, as described in~\cite{huang2023pfb}. 

\Paragraph{User Study.} 
We conduct a user study on the Amazon MTurk platform~\cite{mturk:online} to evaluate our proposed method. 
In each questionnaire, participants are presented with a source image and two edited images: one generated by our proposed method and the other by a randomly selected baseline method. 
The presentation order of the edited images is randomized to avoid bias. 
To enhance their visibility, we outline the desired edited regions with white dashed lines in both the source and edited images.
Additionally, a simplified version of the target edit prompt is displayed beneath the comparison images to facilitate direct comparisons.
Following the evaluation approach of~\cite{magedit}, participants are asked to respond to three questions:
\begin{compactitem}
\item Structure Preservation: In the dashed region, which image preserves structures more similarly to the source image? 
\item Text Alignment: Which image aligns better with the ``edit prompt'' in the dashed region? 
\item Overall: In the dashed region, which image performs better overall?
\end{compactitem}

%
%
%%%%%%%%%%%%%%%%%%%%%%%%%%%%%%%%%%%%%%%%%%%%%%%%%%%%%%%%
\begin{figure*}[!t]
    \centering
    \includegraphics[width=0.98\linewidth]{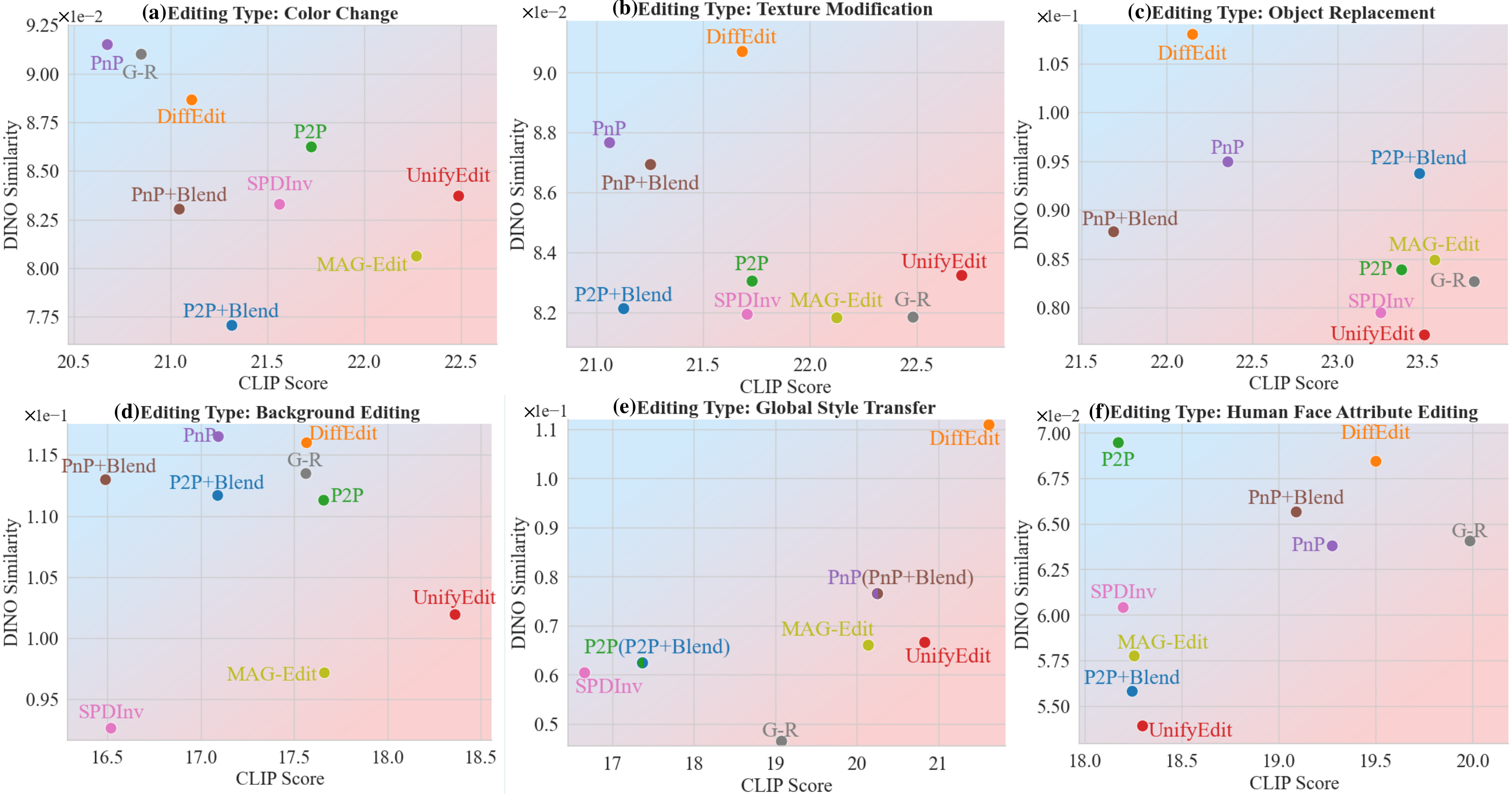}
   % \vspace{-4 mm}
    \caption{
   \textbf{Quantitative comparisons of baselines and our UnifyEdit across various editing types.} 
    We quantify editability and fidelity using CLIP sore (righter is better) and DINO similarity distance (lower is better), respectively.
    Balancing the aspects requires a high CLIP score and relatively low DINO similarity.
    Therefore, points closer to the  \raisebox{0.1ex}{\colorbox[HTML]{fbd0d0}{\strut pink}} region of the background represent better performance, while those closer to the  \raisebox{0.1ex}{\colorbox[HTML]{d0eafc}{\strut blue}} region indicate poorer performance.
    }
    \vspace{-4 mm}
    \label{fig:baseline_quantitative}
\end{figure*}
%%%%%%%%%%%%%%%%%%%%%%%%%%%%%%%%%%%%%%%%%%%%%%%%%%%%%%%%%%%%%%%%%%%%%
%%%%%%%%%%%%%%%%%%%%%%%%%%%%%%%%%%%%%%%%%%%%%%%%
\begin{figure}[!t]
    \centering
    \includegraphics[width=0.99\linewidth]{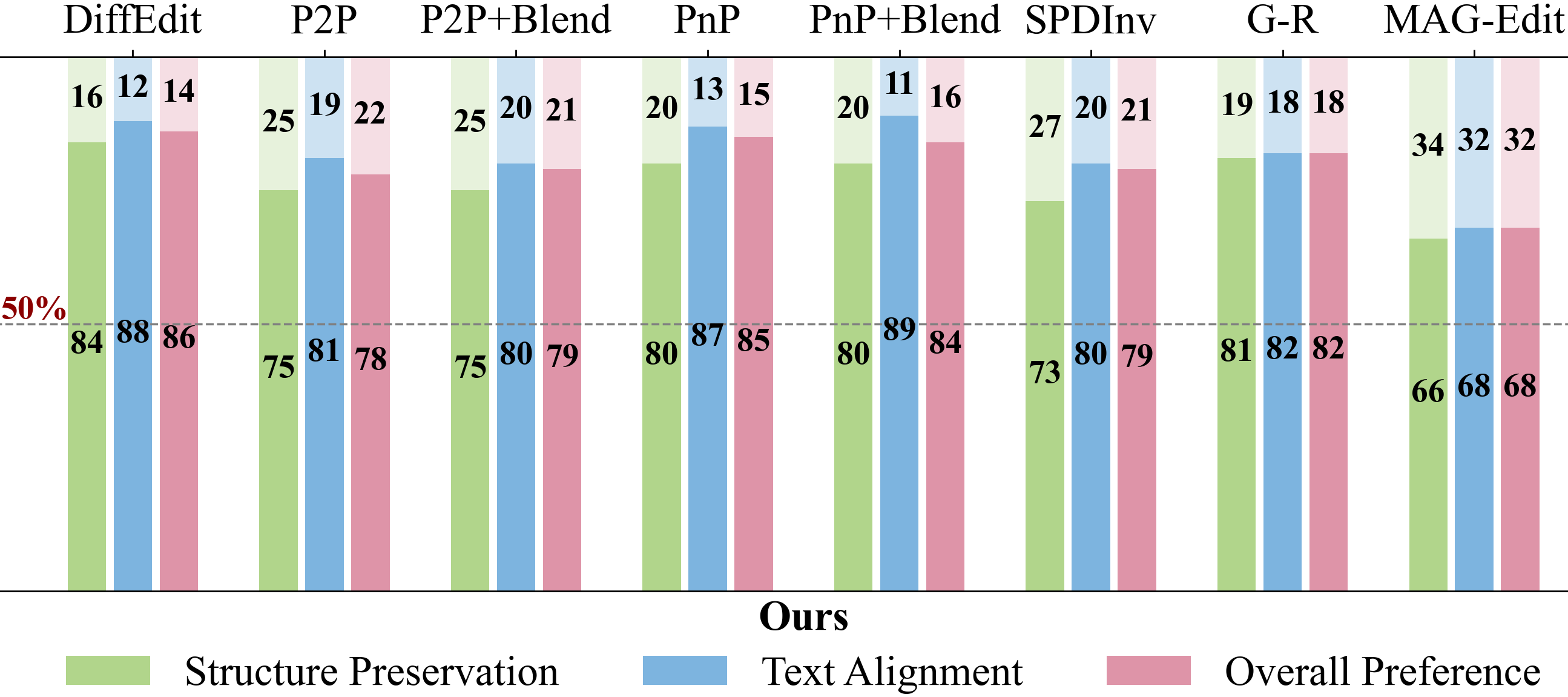} 
    \vspace{-2mm}
    \caption{
    \textbf{Average human preferences across various editing types.}
    The values indicate the proportion of users who preferred our proposed method over comparative approaches.
    }
    \label{fig:user_study}
    \vspace{-1 mm}
\end{figure}
%%%%%%%%%%%%%%%%%%%%%%%%%%%%%%%%%%%%%%%%%%%%%%%%

\subsection{Comparisons with state-of-the-art Methods}
\label{baselines}
\Paragraph{Baselines.} 
We evaluate our approach against existing state-of-the-art TIE methods, categorized as follows:

\begin{compactitem}
\item \textbf{Inpainting-based methods}:
    \textit{DiffEdit}~\cite{couairon2022diffedit} is a typical inpainting method that employs an implicitly predicted mask to preserve the non-editing region.
    
\item \textbf{Attention-injection-based methods}:
    \textit{P2P}~\cite{hertz2022prompt} and \textit{PnP}~\cite{tumanyan2023plug} utilize attention injection to maintain fidelity across the entire image.
    \textit{P2P$+$Blend} and \textit{PnP$+$Blend} are enhanced P2P~\cite{hertz2022prompt} and PnP~\cite{tumanyan2023plug} with blending operations to ensure fidelity outside the editing mask remains unchanged.
\item \textbf{Inversion-based methods}:
    \textit{SPD Inversion (SPD Inv)}~\cite{li2024source} is the latest inversion-based method that focuses on improving the DDIM inversion~\cite{song2020denoising} to achieve more accurate reconstruction results.
\item \textbf{Gradient-based methods}:
    \textit{Guide-and-Rescale (G-R)}~\cite{titov2024guide} is a recent method that leverages noise guidance in TIE.
    Similar to our $\mathcal{L}_{\rm{SAP}}$, it aims at reducing discrepancies between SA maps generated during reconstruction and editing, thereby achieving fidelity.
    \textit{MAG-Edit}~\cite{magedit} builds on attention-injection-based backbones like P2P\cite{hertz2022prompt} to maintain fidelity while introducing two constraints to enhance text alignment for editability.

\end{compactitem}

We use the official codes released by the authors for P2P~\cite{p2p:online}, PnP~\cite{pnp:online},
SPDInv~\cite{spdinv:online},
G-R~\cite{guide_rescale:online}, and
MAG-Edit~\cite{magedit:online}. 
For DiffEdit~\cite{couairon2022diffedit}, we adopt the implementation of InstructEdit~\cite{iedit:online}.
To facilitate fair comparisons, all methods use the \emph{identical masks} provided in our benchmark dataset and the Stable Diffusion v1.4 model as the backbone.
Notably, for DiffEdit~\cite{couairon2022diffedit}, P2P~\cite{hertz2022prompt} + Blend, PnP~\cite{tumanyan2023plug} + Blend and SPDInv~\cite{li2024source}, we utilize ground-truth masks instead of those generated through unsupervised learning or derived from average CA maps.
In the case of P2P~\cite{hertz2022prompt} and MAG-Edit~\cite{magedit}, we also integrate NTI~\cite{mokady2023null} as our approach to encode real images.
%

%%%%%%%%%%%%%%%%%%%%%%%%%%%%%%%%%%%%%%%%
\begin{table}[!t]
\centering
\renewcommand{\arraystretch}{1.1}  % 
\begin{tabular}{p{2.6cm}<{\centering}| p{0.7cm}<{\centering} p{0.7cm}<{\centering} | p{0.7cm}<{\centering} p{0.8cm}<{\centering}  |p{0.7cm}<{\centering}}
\toprule
\multirow{-0.5}{*}{\begin{tabular}[c]{@{}c@{}} \textbf{Method}\end{tabular}} & \textbf{w/o SAP} & \textbf{w/o CAA}   & \multirow{-0.5}{*}{\begin{tabular}[c]{@{}c@{}} \textbf{$\mathcal{G}_{naive}$}\end{tabular}} & \multirow{-0.5}{*}{\begin{tabular}[c]{@{}c@{}} \textbf{$\mathcal{G}_{norm}$}\end{tabular}} & \multirow{-1}{*}{\begin{tabular}[c]{@{}c@{}} \textbf{Unify} \\ \textbf{Edit}\end{tabular}}   \\
\midrule
\rowcolor[HTML]{EFEFEF} 
\multicolumn{6}{c}{\textbf{Color Change}} \\
\midrule
DINO Similarity $\downarrow$
&0.124&	\textbf{0.077} &	0.127 &	0.088 &\underline{0.084}  \\
CLIP Score $\uparrow$
& \underline{21.83} &	20.87 &	21.13 &	21.82 &\textbf{22.49}\\
\midrule
\rowcolor[HTML]{EFEFEF} 
\multicolumn{6}{c}{\textbf{Texture Modification}} \\
\midrule
DINO Similarity $\downarrow$
&0.099 &	\textbf{0.082} &	0.097 &	0.090 &\underline{0.083} \\
CLIP Score $\uparrow$
 &21.48 &	20.39	&19.93	&\underline{22.30} &\textbf{22.71}\\
\midrule
\rowcolor[HTML]{EFEFEF} 
\multicolumn{6}{c}{\textbf{Object Replacement}} \\
\midrule
DINO Similarity $\downarrow$
&0.109 &\underline{0.082} &	0.096 &	0.100  &\textbf{0.077}\\
CLIP Score $\uparrow$
&\underline{23.27} &	22.76 &	22.78&	23.16 &\textbf{23.51}\\
\midrule
\rowcolor[HTML]{EFEFEF} 
\multicolumn{6}{c}{\textbf{Background Editing}} \\
\midrule
DINO Similarity $\downarrow$
&0.146  &\underline{0.104}    &0.157   &0.122 &\textbf{0.102}\\
CLIP Score $\uparrow$
&18.33  &16.36    &16.67  &\textbf{18.88} &\underline{18.36}\\
\midrule
\rowcolor[HTML]{EFEFEF} 
\multicolumn{6}{c}{\textbf{Global Style Transfer}} \\
\midrule
DINO Similarity $\downarrow$
&0.113  &\textbf{0.053}    &0.091  &0.081   &\underline{0.066}\\
CLIP Score $\uparrow$
&\underline{22.77}  &17.24   &17.19  &\textbf{23.04} &20.83\\
\midrule
\rowcolor[HTML]{EFEFEF} 
\multicolumn{6}{c}{\textbf{Human Face Attribute Editing}} \\
\midrule
DINO Similarity $\downarrow$
&0.083	&\textbf{0.053}	&0.099 &	0.071 &\underline{0.054}\\
CLIP Score $\uparrow$
&\underline{19.53} &	17.75&	18.89	&\textbf{19.55} &18.30\\
\bottomrule
\end{tabular}
\caption{\textbf{Quantitative results of ablation study.}
Bold and underline indicate the best and second best value, respectively.}
\label{tab:self_constraint}
%\vspace{-1 mm}
\end{table}
%%%%%%%%%%%%%%%%%%%%%%%%%%%%%%%%%%%%%%%%%%%%%%%%%%%

\Paragraph{Qualitative Results.}
\figref{qualitative} shows that DiffEdit~\cite{couairon2022diffedit}, which employs DDIM inversion~\cite{song2020denoising} for foreground generation, significantly alters the structure fidelity across all editing types.
For example, while it successfully generates an ``angry woman", it loses critical identity information from the source image.
Attention-injection-based methods like P2P~\cite{hertz2022prompt} and PnP~\cite{tumanyan2023plug} effectively preserve the original structure in editing scenarios that do not require shape variations. 
However, as discussed in~\subsecref{motivation}, directly copying attention maps restricts flexibility in editability, leading to under-editing issues such as insufficient color change and texture modification.
As shown in the first four rows of \figref{qualitative}, these methods fail to generate the corresponding colors and textures specified by the text prompts. On the other hand, for significant shape variations, such as changing a ``dog"  to an ``orangutan," these methods may alter the posture too drastically compared to the source image, leading to over-editing issues.
Combined with the blending operation, P2P$+$Blend and PnP$+$Blend still exhibit the same issues, although they effectively preserve the regions outside the mask.
While SPDInv~\cite{li2024source} enhances DDIM inversion to generate more accurate features, its reliance on attention injection still limits its effectiveness in achieving a balanced outcome.
The gradient-based method, G-R~\cite{titov2024guide}, which optimizes noise $\epsilon^\ast_\theta$ to minimize the gap between $A^{self}$ and $A^{*self}$, demonstrates greater flexibility in structure preservation compared to attention-injection-based methods, particularly in human face attribute editing.
However, due to its limited focus on text alignment, its editing effects are constrained in scenarios that require higher editability, as demonstrated in examples such as the ``cartoon" style.
Although MAG-Edit~\cite{magedit} focuses on enhancing text alignment, its reliance on attention injection to preserve fidelity limits its editability, resulting in under-editing issues in cases like the ``purple pear".
In contrast, our method demonstrates superior editing adaptability, effectively balancing fidelity and editability across a wide range of editing types.

%%%%%%%%%%%%%%%%%%%%%%%%%%%%%%%%%%%%%%%%%%%%%%%%
\begin{figure}[!t]
    \centering
    \includegraphics[width=0.99\linewidth]{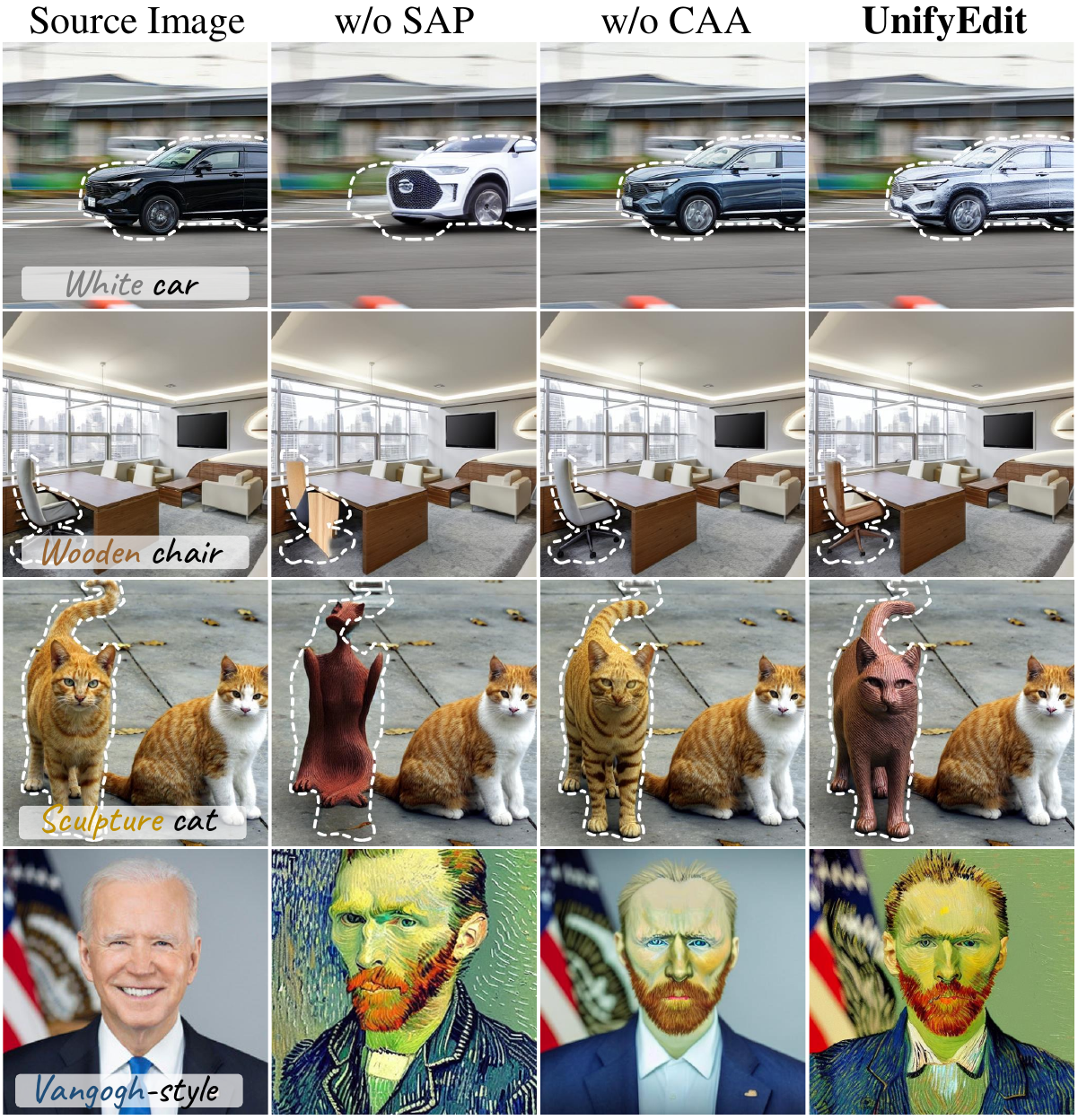} 
    \vspace{-2 mm}
    \caption{
    \textbf{Qualitative results of ablation study on attention-based constraints.}
    White dashed outlines are used to highlight the target object in foreground editing.
   Combining both terms is crucial for achieving a good balance between fidelity and editability.
    }
    \label{fig:self_constraint}
    \vspace{-1 mm}
\end{figure}
%%%%%%%%%%%%%%%%%%%%%%%%%%%%%%%%%%%%%%%%%%%%%%%%
%%%%%%%%%%%%%%%%%%%%%%%%%%%%%%%%%%%%%%%%%%
\begin{figure}[!t]
    \centering
    \includegraphics[width=0.99\linewidth]{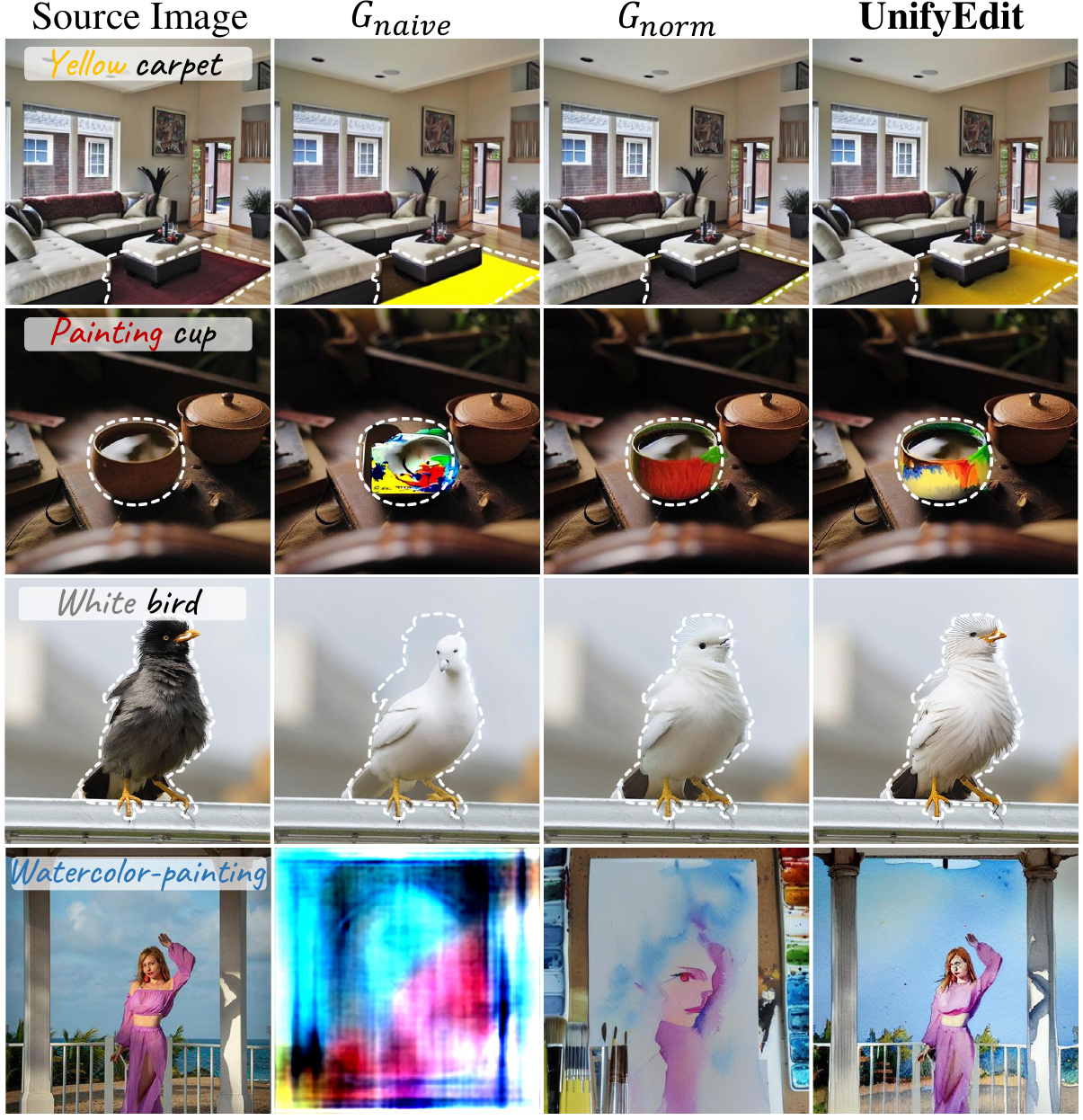} 
    \vspace{-2 mm}
    \caption{
    \textbf{Qualitative results of ablation study on different gradients.}
    The target object is accentuated with white dashed outlines in the foreground editing.
    $\mathcal{G}_{naive}$
    in~\eqnref{a2} can lead to over-editing and, in some cases, image collapse. 
    While $\mathcal{G}_{norm}$ in \eqnref{a3} mitigates these issues, it still encounters both under-editing and over-editing failures. 
    In contrast, our method, which employs $\mathcal{G}_{blc}$ in 
 \eqnref{a5}, successfully achieves a balanced result.
    }
    \label{fig:atb}
    %\vspace{-2 mm}
\end{figure}
%%%%%%%%%%%%%%%%%%%%%%%%%%%%%%%%%%%%%%%%%%

%HERE
\Paragraph{Quantitative Results.}
\figref{baseline_quantitative} shows quantitative results of the evaluation methods with 
CLIP score on the x-axis and DINO similarity on the y-axis.
The points on the bottom-right  (high CLIP score and low DINO similarity) represent better balance performance. 
%Conversely, points located towards the top-left, characterized by a low CLIP score and high DINO similarity, signify poorer balance performance.
%
We use a colormap to visualize the performance of the compared methods and ours, ranging from \raisebox{0.0ex}{\colorbox[HTML]{d0eafc}{\strut blue}} to \raisebox{0.0ex}{\colorbox[HTML]{fbd0d0}{\strut pink}}.
As shown in~\figref{baseline_quantitative}(a)(b)(d), our method performs favorably against the baselines by achieving better alignment with the target text for edits involving minimal shape variations, such as those related to color and texture. 
Although object replacement and global style transfer entail significant shape or texture changes, maintaining structural consistency is vital to preserving the source image's visual integrity. 
Our approach records the lowest DINO similarity for human facial attribute editing, as shown in \figref{baseline_quantitative}(f). 
While DiffEdit~\cite{couairon2022diffedit} and G-R~\cite{titov2024guide} obtain higher CLIP scores, these methods do not preserve the subject's identity, as demonstrated in the last two rows of \figref{qualitative}.
These results show that our method performs well with a robust balance compared to other baselines across various editing types.

%%%%%%%%%%%%%%%%%%%%%%%%%%%%%%%%%%%%%%%%%%%%%%%%%
%\newpage
\begin{figure*}[!t]
    \centering
    \includegraphics[width=0.98\linewidth]{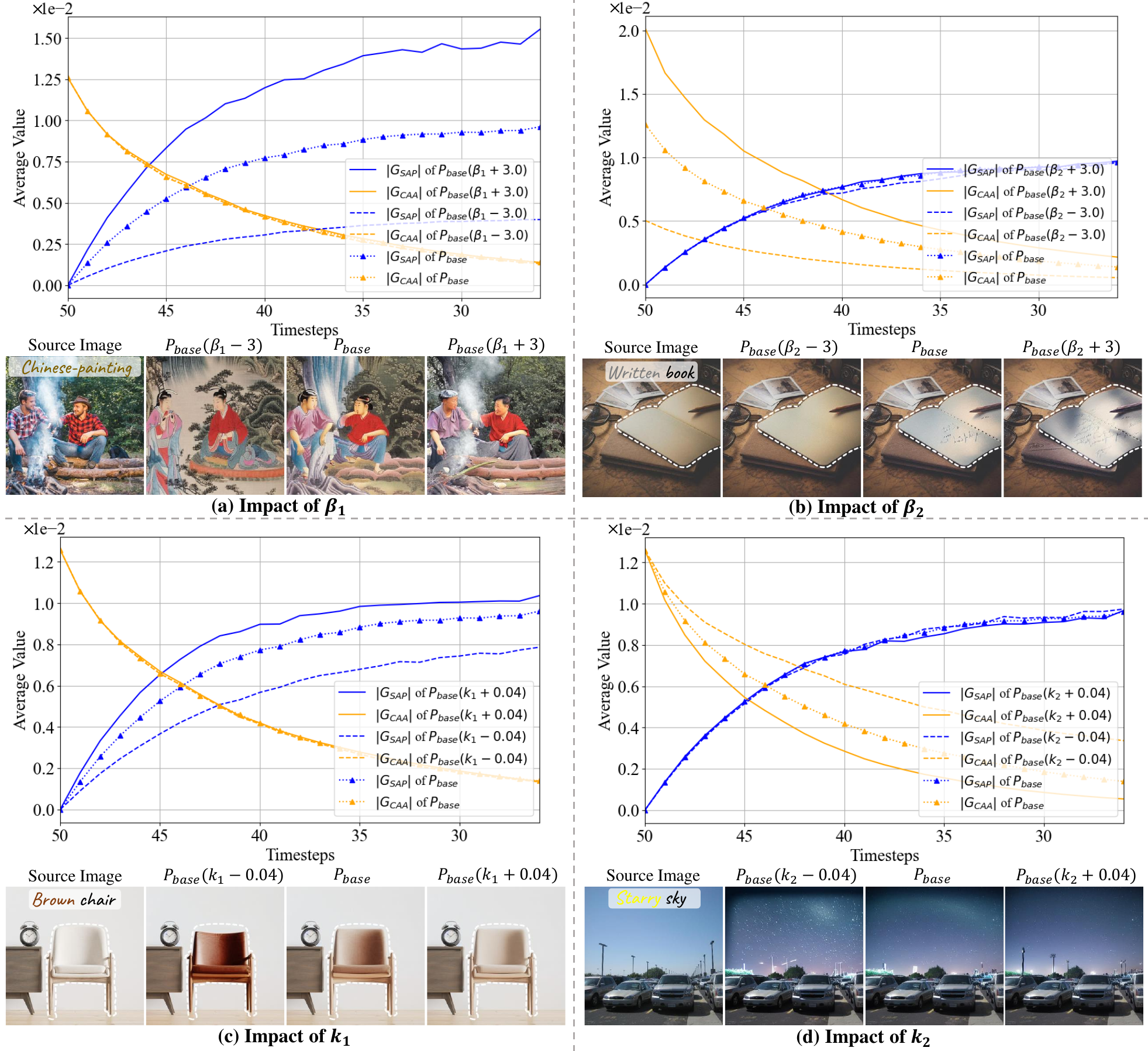}    
    \caption{
   \textbf{Ablation study on hyper-parameters in adaptive time-step scheduler.}
    The scaling factors $\beta_1$ and $\beta_2$, along with the rate factors $k_1$ and $k_2$, regulate the magnitude and changing rate, influencing the editing outcomes. 
   }
    %\vspace{-3 mm}
    \label{fig:parameters}
\end{figure*}
%%%%%%%%%%%%%%%%%%%%%%%%%%%%
\Paragraph{User Study.}
To ensure reliability, we invite Amazon MTurk workers with `Master' status and a Human Intelligence Task (HIT) Approval Rate exceeding $90\%$ across all Requesters' HITs.
We collect 1,750 completed questionnaires from these subjects.
As shown in \figref{user_study}, the percentages indicate the proportion of participants who preferred our proposed method over baseline approaches.
For fidelity, a significant majority, ranging from $66\%$ to $84\%$, indicates that our method demonstrates superior structure preservation compared to existing methods.
Regarding editability, our method is preferred for improved text alignment, with preference rates ranging from $68\%$ to $89\%$.
Overall, our proposed method is favored by 68\% to 86\% 
of participants due to its effective balance between editability and fidelity.

%HERE
\subsection{Ablation Study on Attention-Based Constraints}
\label{ablation1}
We conduct ablation studies on the following variations to validate the role of two attention-based constraints,
\begin{enumerate}
\item \textbf{w/o SAP}: diffusion latent feature $z_t$ is optimized without the gradient of $\mathcal{L}_{\rm{SAP}}$, meaning that the $\mathcal{G}_{blc}$ in \eqnref{a5} is replaced with $\textcolor{orange}{\mathcal{G}_{\rm{CAA}}} = \lambda_2^\ast\frac{\nabla_{z_t^*}\mathcal{L}_{\rm{CAA}}}{||\nabla_{z_t^*}\mathcal{L}_{\rm{CAA}}||_2}$.
\item \textbf{w/o CAA}: diffusion latent feature $z_t$ is optimized without the gradient of $\mathcal{L}_{\rm{CAA}}$, so $\mathcal{G}_{blc}$ in \eqnref{a5} is replaced with $\textcolor{blue}{\mathcal{G}_{\rm{SAP}}}  = \lambda_1^\ast\frac{\nabla_{z_t^*}\mathcal{L}_{\rm{SAP}}}{||\nabla_{z_t^*}\mathcal{L}_{\rm{SAP}}||_2}$.
\end{enumerate}
As shown in \figref{self_constraint}, the absence of $\mathcal{L}_{\rm{SAP}}$ results in strong editing effects that align closely with the prompt but introduce significant structural discrepancies.
For example, both the ``wooden chair" lose structural integrity in the \textbf{w/o SAP} examples, causing noticeable artifacts. 
This is reflected in a relatively high CLIP score but a very low DINO similarity, as presented in \tabref{self_constraint}.
In contrast, structural fidelity is maintained without the influence of $\mathcal{L}_{\rm{CAA}}$, but the lack of sufficient text alignment results in unsatisfactory edits. 
For instance, the ``sculpture cat" retains excessive structure information from the source images, compromising texture changes' editability. 
As a result, both the CLIP score and DINO similarity are low, as presented in \tabref{self_constraint}.
As shown in the last column of \figref{self_constraint}, utilizing both constraints together yields the best results.

% 
%%%%%%%%%%%%%%%%%%%%%%%%%%%%%%%%%%%%%%%%%%%%%%%%%%%%%%%%
\begin{figure}[!t]
    \centering
    \includegraphics[width=0.95\linewidth]{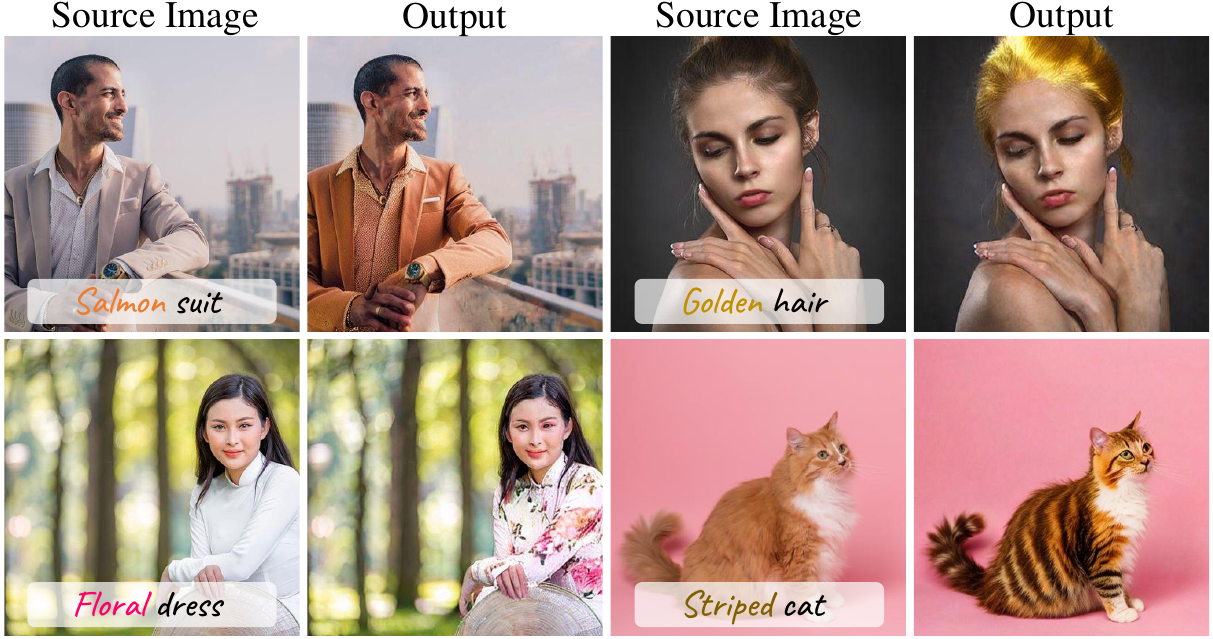}
    \caption{
   \textbf{Editing results using DDIM inversion.}
The proposed method maintains effectiveness by employing SA constraints derived from the SA maps generated during the DDIM inversion.
   }
    %\vspace{-2 mm}
    \label{fig:fatezero}
\end{figure}
%%%%%%%%%%%%%%%%%%%%%%%%%%%%%%%%%%%%%%%%%%
%%%%%%%%%%%%%%%%%%%%%%%%%%%%%%%%%%%%%%%%%%%%%%%%%%%%%%%%
\begin{figure}[!t]
    \centering
    \includegraphics[width=0.95\linewidth]{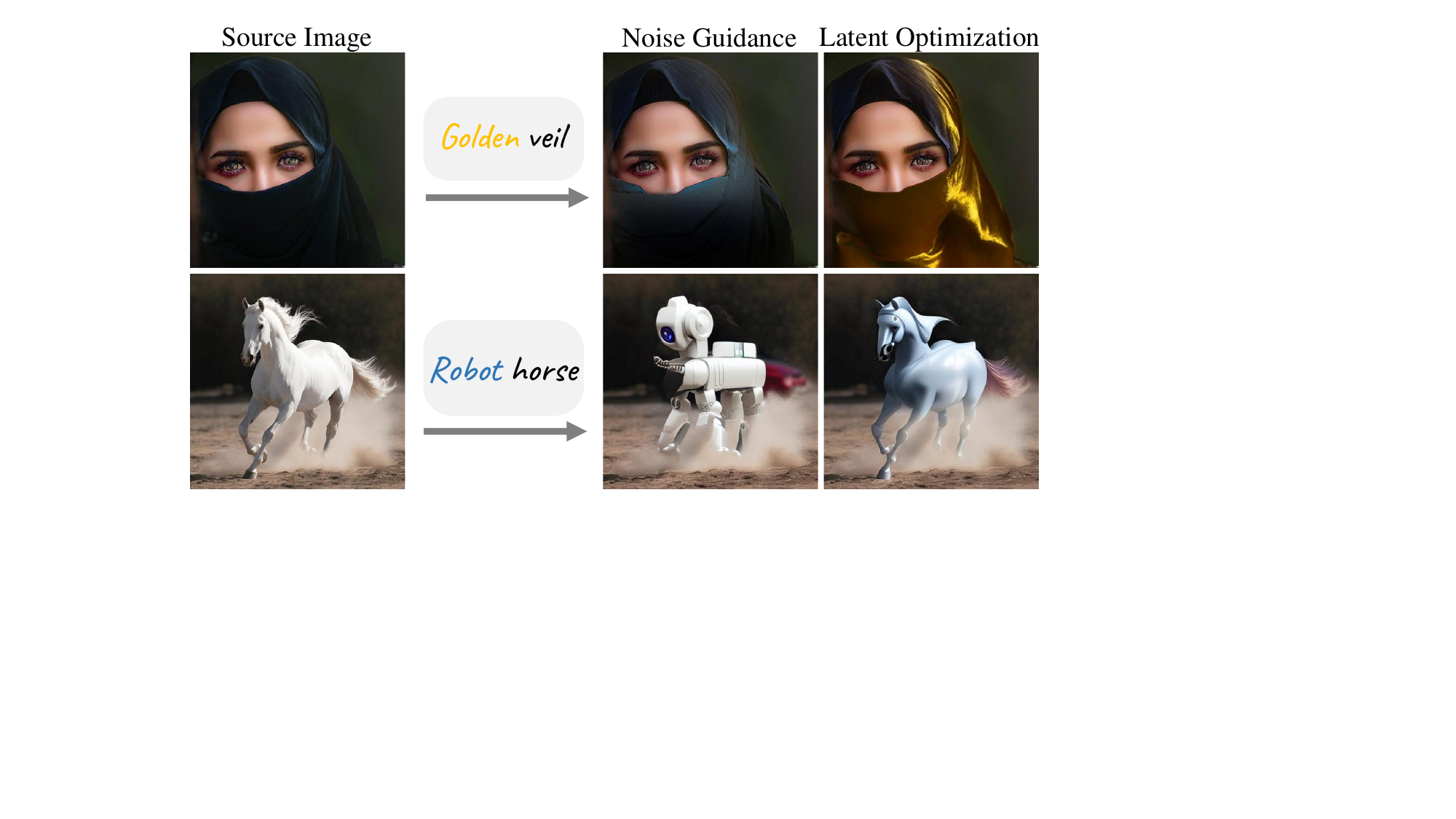}
    \caption{
   \textbf{Diffusion latent optimization \vs noise guidance.} 
Latent optimization outperforms noise guidance in balancing fidelity and editability.
   }
   % \vspace{-1 mm}
    \label{fig:noise_guidance}
\end{figure}
%%%%%%%%%%%%%%%%%%%%%%%%%%%%%%%%%%%%%%%%%%

\subsection{Ablation Study on Adaptive Time-Step Scheduler}
We conduct ablation studies from two perspectives
to explore the effectiveness of the adaptive time-step scheduler:

\Paragraph{Impact on the Different Gradients.}
We compare with optimization using  $\mathcal{G}_{naive}$ in \eqnref{a2},  $\mathcal{G}_{norm}$ in \eqnref{a3} and $\mathcal{G}_{blc}$ in \eqnref{a5} (\ie UnifyEdit), respectively.
As discussed in \subsecref{atb}, the direct combination of $\mathcal{L}_{\rm{SAP}}$ and $\mathcal{L}_{\rm{CAA}}$ for optimizing $z_t$ results in  the predominance of 
$\mathcal{L}_{\rm{CAA}}$'s gradient.
This dominance can lead to a significant loss of structural fidelity and image collapse demonstrated in~\figref{atb}.
Utilizing the $L_2$ norm to balance the two constraints mitigates the risk of image collapse, yet it still results in over- or under-editing issues.
As shown in~\figref{atb}, the ``painting cup" exhibits weak editing effects, while the structure of the ``white bird" deviates from the original image.
In contrast, the proposed adaptive time-step scheduler effectively balances the influence of the two constraints, achieving optimal editing results across a diverse range of editing scenarios.
%
%%%%%%%%%%%%%%%%%%%%%%%%%%%%%%%%%%%%%%%%%
\begin{table}[!t]
\setlength{\extrarowheight}{2pt} % 每行增加 2pt 的高度
\centering
\footnotesize
\begin{tabular}{c@{\hspace{6pt}}c@{\hspace{6pt}}c@{\hspace{8pt}}c@{\hspace{7pt}}c@{\hspace{6pt}}c@{\hspace{6pt}}c@{\hspace{6pt}}c@{\hspace{7pt}}c@{\hspace{3pt}}}
\toprule
\rowcolor[HTML]{EFEFEF} 
Method 
& \begin{tabular}[c]{@{}c@{}}DiffEdit\\ \cite{couairon2022diffedit}\end{tabular} 
&\begin{tabular}[c]{@{}c@{}}P2P\\ \cite{hertz2022prompt}\end{tabular}   
& \begin{tabular}[c]{@{}c@{}}PnP\\ \cite{tumanyan2023plug}\end{tabular} 
& \begin{tabular}[c]{@{}c@{}}SPDInv\\ \cite{li2024source}\end{tabular}  
& \begin{tabular}[c]{@{}c@{}}G-R\\  \cite{titov2024guide}\end{tabular}  
& \begin{tabular}[c]{@{}c@{}}MAG-\\ Edit\cite{magedit}\end{tabular}  
& \begin{tabular}[c]{@{}c@{}}\textbf{Ours}$+$\\NTI\cite{mokady2023null}\end{tabular}  \\
\midrule
\begin{tabular}[c]{@{}c@{}}Inversion \\Time (s)\end{tabular}    
&4.2  &87.9  &46.5  &21.5    &3.0      &87.9   &87.9  \\
\rowcolor[HTML]{EFEFEF} 
\begin{tabular}[c]{@{}c@{}}Denoising \\Time (s)\end{tabular}   
&5.3  &10.7  &93.2  &10.6    &30.2      &83.9  &24.2 \\
\begin{tabular}[c]{@{}c@{}}Memory\\ (GB)\end{tabular} &10.8  &12.8  &17.6  &16.2  &25.7   &19.5 &21.4\\
\bottomrule
\end{tabular}
\caption{\textbf{Runtime and GPU memory requirements for the baselines and our proposed method.}}
\label{tab:memory}
\vspace{-1 mm}
\end{table}
%%%%%%%%%%%%%%%%%%%%%%%%%%%%%%%%%%%%%%%%%%%%%%%%%%% 

\Paragraph{Impact on the Hyper-Parameters.}
As discussed in \subsecref{atb}, the scaling factors $\beta_1$, $\beta_2$, and rate factors $k_1$, $k_2$ are essential for adjusting the weights of $\mathcal{L}_{\rm{SAP}}$ and $\mathcal{L}_{\rm{CAA}}$ in the adaptive time-step scheduler.
As outlined in \subsecref{implementation_details}, 
we define the standard parameter settings for various editing tasks as baseline parameters, represented as $P_{base} = (\beta_1, \beta_2, k_1, k_2)_j$,
where $j$ refers to a specific editing type.
To investigate the impact of $\beta_1$, we adjust its baseline value by adding or subtracting $3.0$,  denoted as $P_{base}(\beta_1 \pm 3)$, while maintaining the other parameters constant.
Similarly, for the rate factors, we modulate $k_1$ by modifying its baseline value by $\pm 0.04$, expressed as $P_{base}(k_1 \pm 0.04)$.
The same procedure is applied to $\beta_2$ and $k_2$.
We visualize the mean absolute value of \textcolor{blue}{$\mathcal{G}_{\rm{SAP}}$} and \textcolor{orange}{$\mathcal{G}_{\rm{CAA}}$} over timesteps under different settings across various editing types.
The gradient variations induced by the parameters and the resulting differences in editing effects are consistent across these types.
We present the average values across all editing types in~\figref{parameters} for simplicity.
As illustrated in~\figref{parameters}(a)(b), the scaling factors determine the overall magnitude of the gradients, represented by the vertical shift of the curve.
For instance, $P_{base}(\beta_1 + 3)$ raises the \textcolor{blue}{$\mathcal{G}_{\rm{SAP}}$} curve, causing the final gradient guidance on $z_t$ to shift away from $\mathcal{L}_{\rm{CAA}}$.
As shown in~\figref{parameters}(a), this results in weaker style rendering for the ``Chinese painting'' task.
Conversely, $P_{base}(\beta_1 - 3)$ lowers the \textcolor{blue}{$\mathcal{G}_{\rm{SAP}}$} curve, leading to weaker editing effects. 
Notably, $\beta_2$ has a similar influence on the \textcolor{orange}{$\mathcal{G}_{\rm{CAA}}$} curve, but its effect leads to the opposite outcome in the final edited results. 
The editing effects for the ``written" token are prominent in $P_{base}(\beta_2 + 3)$, while they are negligible in $P_{base}(\beta_2 - 3)$ (see~\figref{parameters}(a)). 
The rate factors affect the rate at which the gradients change, reflected in the steepness of the curves, as shown in \figref{parameters}(c)(d).
For instance, $P_{base}(k_1 + 0.04)$ causes the \textcolor{blue}{$\mathcal{G}_{\rm{SAP}}$} curve to rise more quickly, increasing the influence of $\mathcal{L}_{\rm{SAP}}$ on the latent $z_t$.
In contrast, a smaller $P_{base}(k_1 - 0.04)$ slows the increase of the \textcolor{blue}{$\mathcal{G}_{\rm{SAP}}$} curve, resulting a higher influence of $\mathcal{L}_{\rm{CAA}}$.
The editing effects for the ``brown chair" are stronger under $P_{base}(k_1 - 0.04)$ and weaker under $P_{base}(k_1 + 0.04)$ (see~\figref{parameters}(c)).
Similarly, $k_2$ has the same impact on the descent rate of \textcolor{orange}{$\mathcal{G}_{\rm{CAA}}$} and the overall editing results.
%

%%%%%%%%%%%%%%%%%%%%%%%%%%%%%%%%%%%%%%%%%%%%%%%%%%%%%%%%
\begin{figure*}[!t]
    \centering
    \includegraphics[width=0.99\linewidth]{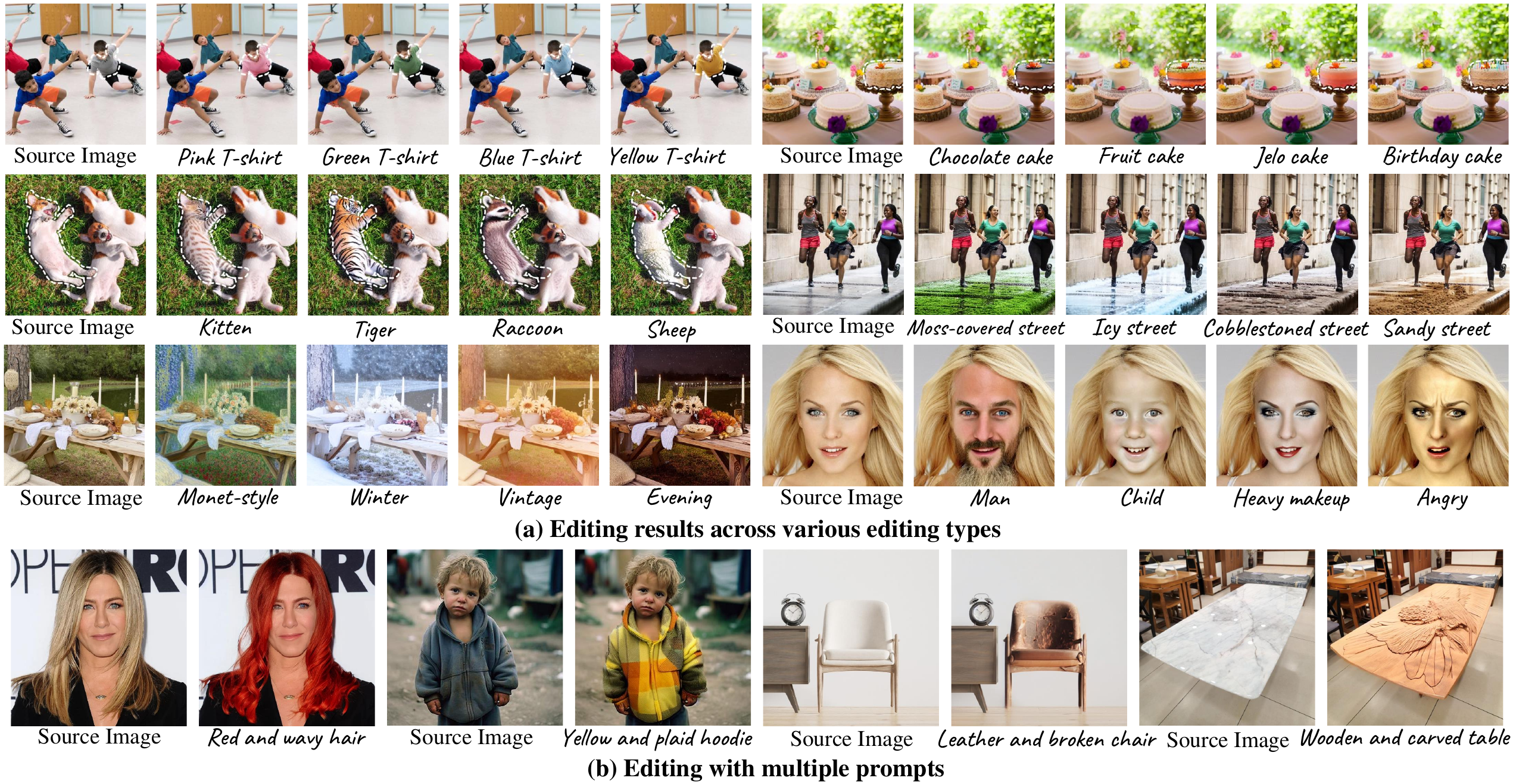}    
    \vspace{-2 mm}
    \caption{
   \textbf{More editing results of UnifyEdit.}   
   We highlight the target object with white dashed outlines in foreground editing.
   UnifyEdit can achieve balance across various editing types and can be applied to multiple target editing tokens.
   }
    \vspace{-5 mm}
    \label{fig:results}
\end{figure*}
%%%%%%%%%%%%%%%%%%%%%%%%%%%%

\subsection{Discussions}
\label{subsec:discussion}

\Paragraph{Compatibility with Other Inversion Methods.}
As discussed in~\subsecref{self_constraints}, our proposed method is compatible with other inversion techniques.
We demonstrate an extreme case aimed at minimizing the gap between the SA maps from the target branch and those generated during the DDIM inversion~\cite{song2020denoising} process defined in~\eqnref{inversion}, similar to the inversion attention fusion proposed in~\cite{qi2023fatezero}.
\figref{fatezero} shows that our method integrates successfully with this approach, producing the desired editing results.

\Paragraph{Comparisons with Noise Guidance.}
As discussed in~\subsecref{latent}, latent optimization uses the gradient $g$ to directly optimize $z_t$, resulting in $\hat{z}_t = z_t - g$.
Instead, noise guidance updates $z_{t-1}$ by using the gradient to adjust the noise estimate, yielding $\hat{\epsilon}^t_\theta = \epsilon^t_\theta - g$. 
Using noise guidance results in the ineffective ``golden veil'' and the loss of structural integrity in the ``robot horse'', demonstrating its relative ineffectiveness in balancing editability and fidelity compared to our method (see~\figref{noise_guidance}).

\Paragraph{Runtime and Memory Usage.}
We report the runtime and GPU memory usage for our proposed method with NTI~\cite{mokady2023null} and the baseline methods on an Nvidia A100 (40GB) GPU in~\tabref{memory}.
The time consumption and memory usage of our method are mainly attributed to latent optimization in the denoising process, yet they remain moderate compared to the other baselines.

\Paragraph{Additional Results.}
As shown in \figref{results}(a), our method effectively balances fidelity and editability across various editing tasks.
Furthermore, as demonstrated in \figref{results}(b), our proposed method can also be applied to multiple target tokens (\eg ``red and wavy").

\section{Conclusions and Future works}
\label{sec:conclusions}
In this work, we present one of the initial efforts to explicitly model the balance between fidelity and editability within a unified diffusion latent optimization framework. 
Our approach is novel in two ways: It incorporates attention-based constraints from the SA and CA that control fidelity and editability, and an adaptive time-step scheduler that balances these constraints.
Quantitative and qualitative results demonstrate that UnifyEdit achieves a superior balance against existing methods across a broad spectrum of editing tasks. 
Overall, our method significantly advances the field of tuning-free diffusion-based TIE by offering a unified approach that explicitly controls the balance between fidelity and editability.
This approach not only meets diverse editing requirements but can also be adjusted dynamically to align with users' preferences.

However, since the SA map captures extensive layouts and semantic information, the proposed SA preservation constraint somewhat constrains the rigidity of target objects. Consequently, our method may face challenges with non-rigid transformations, such as changing a sitting dog into a jumping dog.
We aim to address these challenges by developing a non-rigid self-attention constraint to enhance the method's adaptability to dynamic transformations in future work.

%\vfill
%\clearpage
\bibliographystyle{IEEEtran}
\bibliography{reference}

\begin{thebibliography}{10}
\providecommand{\url}[1]{#1}
\csname url@samestyle\endcsname
\providecommand{\newblock}{\relax}
\providecommand{\bibinfo}[2]{#2}
\providecommand{\BIBentrySTDinterwordspacing}{\spaceskip=0pt\relax}
\providecommand{\BIBentryALTinterwordstretchfactor}{4}
\providecommand{\BIBentryALTinterwordspacing}{\spaceskip=\fontdimen2\font plus
\BIBentryALTinterwordstretchfactor\fontdimen3\font minus \fontdimen4\font\relax}
\providecommand{\BIBforeignlanguage}[2]{{%
\expandafter\ifx\csname l@#1\endcsname\relax
\typeout{** WARNING: IEEEtran.bst: No hyphenation pattern has been}%
\typeout{** loaded for the language `#1'. Using the pattern for}%
\typeout{** the default language instead.}%
\else
\language=\csname l@#1\endcsname
\fi
#2}}
\providecommand{\BIBdecl}{\relax}
\BIBdecl

\bibitem{ramesh2022hierarchical}
A.~Ramesh, P.~Dhariwal, A.~Nichol, C.~Chu, and M.~Chen, ``Hierarchical text-conditional image generation with clip latents,'' \emph{arXiv preprint arXiv:2204.06125}, 2022.

\bibitem{rombach2022high}
R.~Rombach, A.~Blattmann, D.~Lorenz, P.~Esser, and B.~Ommer, ``High-resolution image synthesis with latent diffusion models,'' in \emph{CVPR}, 2022.

\bibitem{saharia2022photorealistic}
C.~Saharia, W.~Chan, S.~Saxena, L.~Li, J.~Whang, E.~L. Denton, K.~Ghasemipour, R.~Gontijo~Lopes, B.~Karagol~Ayan, T.~Salimans \emph{et~al.}, ``Photorealistic text-to-image diffusion models with deep language understanding,'' in \emph{NeurIPS}, 2022.

\bibitem{brooks2023instructpix2pix}
T.~Brooks, A.~Holynski, and A.~A. Efros, ``Instructpix2pix: Learning to follow image editing instructions,'' in \emph{CVPR}, 2023.

\bibitem{zhang2023magicbrush}
K.~Zhang, L.~Mo, W.~Chen, H.~Sun, and Y.~Su, ``Magicbrush: A manually annotated dataset for instruction-guided image editing,'' in \emph{NeurIPS}, 2023.

\bibitem{kawar2023imagic}
B.~Kawar, S.~Zada, O.~Lang, O.~Tov, H.~Chang, T.~Dekel, I.~Mosseri, and M.~Irani, ``Imagic: Text-based real image editing with diffusion models,'' in \emph{CVPR}, 2023.

\bibitem{zhang2023sine}
Z.~Zhang, L.~Han, A.~Ghosh, D.~N. Metaxas, and J.~Ren, ``Sine: Single image editing with text-to-image diffusion models,'' in \emph{CVPR}, 2023.

\bibitem{hertz2022prompt}
A.~Hertz, R.~Mokady, J.~Tenenbaum, K.~Aberman, Y.~Pritch, and D.~Cohen-Or, ``Prompt-to-prompt image editing with cross attention control,'' in \emph{ICLR}, 2023.

\bibitem{tumanyan2023plug}
N.~Tumanyan, M.~Geyer, S.~Bagon, and T.~Dekel, ``Plug-and-play diffusion features for text-driven image-to-image translation,'' in \emph{CVPR}, 2023.

\bibitem{parmar2023zero}
G.~Parmar, K.~Kumar~Singh, R.~Zhang, Y.~Li, J.~Lu, and J.-Y. Zhu, ``Zero-shot image-to-image translation,'' in \emph{SIGGRAPH}, 2023.

\bibitem{mokady2023null}
R.~Mokady, A.~Hertz, K.~Aberman, Y.~Pritch, and D.~Cohen-Or, ``Null-text inversion for editing real images using guided diffusion models,'' in \emph{CVPR}, 2023.

\bibitem{cao_2023_masactrl}
M.~Cao, X.~Wang, Z.~Qi, Y.~Shan, X.~Qie, and Y.~Zheng, ``Masactrl: Tuning-free mutual self-attention control for consistent image synthesis and editing,'' in \emph{ICCV}, 2023.

\bibitem{couairon2022diffedit}
G.~Couairon, J.~Verbeek, H.~Schwenk, and M.~Cord, ``Diffedit: Diffusion-based semantic image editing with mask guidance,'' in \emph{ICLR}, 2023.

\bibitem{magedit}
Q.~Mao, L.~Chen, Y.~Gu, Z.~Fang, and M.~Z. Shou, ``Mag-edit: Localized image editing in complex scenarios via mask-based attention-adjusted guidance,'' in \emph{ACM MM}, 2024.

\bibitem{avrahami2023blended}
O.~Avrahami, O.~Fried, and D.~Lischinski, ``Blended latent diffusion,'' \emph{ACM TOG}, 2023.

\bibitem{qiao2024baret}
Y.~Qiao, F.~Wang, J.~Su, Y.~Zhang, Y.~Yu, S.~Wu, and G.-J. Qi, ``Baret: Balanced attention based real image editing driven by target-text inversion,'' in \emph{AAAI}, 2024.

\bibitem{li2024source}
R.~Li, R.~Li, S.~Guo, and L.~Zhang, ``Source prompt disentangled inversion for boosting image editability with diffusion models,'' in \emph{ECCV}, 2024.

\bibitem{titov2024guide}
V.~Titov, M.~Khalmatova, A.~Ivanova, D.~Vetrov, and A.~Alanov, ``Guide-and-rescale: Self-guidance mechanism for effective tuning-free real image editing,'' in \emph{ECCV}, 2024.

\bibitem{song2020denoising}
J.~Song, C.~Meng, and S.~Ermon, ``Denoising diffusion implicit models,'' in \emph{ICLR}, 2021.

\bibitem{ju2023direct}
X.~Ju, A.~Zeng, Y.~Bian, S.~Liu, and Q.~Xu, ``Pnp inversion: Boosting diffusion-based editing with 3 lines of code,'' in \emph{ICLR}, 2024.

\bibitem{liu2024towards}
B.~Liu, C.~Wang, T.~Cao, K.~Jia, and J.~Huang, ``Towards understanding cross and self-attention in stable diffusion for text-guided image editing,'' in \emph{CVPR}, 2024.

\bibitem{ho2020denoising}
J.~Ho, A.~Jain, and P.~Abbeel, ``Denoising diffusion probabilistic models,'' in \emph{NeurIPS}, 2020.

\bibitem{Kim_diffusionclip}
G.~Kim, T.~Kwon, and J.~C. Ye, ``Diffusionclip: Text-guided diffusion models for robust image manipulation,'' in \emph{CVPR}, 2022.

\bibitem{MingiAsyrp}
M.~Kwon, J.~Jeong, and Y.~Uh, ``Diffusion models already have {A} semantic latent space,'' in \emph{ICLR}, 2023.

\bibitem{ValevskiUniTune}
D.~Valevski, M.~Kalman, E.~Molad, E.~Segalis, Y.~Matias, and Y.~Leviathan, ``Unitune: Text-driven image editing by fine tuning a diffusion model on a single image,'' \emph{ACM TOG}, 2022.

\bibitem{brack2024ledits}
M.~Brack, F.~Friedrich, K.~Kornmeier, L.~Tsaban, P.~Schramowski, K.~Kersting, and A.~Passos, ``Ledits++: Limitless image editing using text-to-image models,'' in \emph{CVPR}, 2024.

\bibitem{guo2023focus}
Q.~Guo and T.~Lin, ``Focus on your instruction: Fine-grained and multi-instruction image editing by attention modulation,'' in \emph{CVPR}, 2024.

\bibitem{zhang2024hive}
S.~Zhang, X.~Yang, Y.~Feng, C.~Qin, C.-C. Chen, N.~Yu, Z.~Chen, H.~Wang, S.~Savarese, S.~Ermon \emph{et~al.}, ``Hive: Harnessing human feedback for instructional visual editing,'' in \emph{CVPR}, 2024.

\bibitem{dhariwal2021diffusion}
P.~Dhariwal and A.~Nichol, ``Diffusion models beat gans on image synthesis,'' in \emph{NeurIPS}, 2021.

\bibitem{avrahami2022blended}
O.~Avrahami, D.~Lischinski, and O.~Fried, ``Blended diffusion for text-driven editing of natural images,'' in \emph{CVPR}, 2022.

\bibitem{huang2023pfb}
W.~Huang, S.~Tu, and L.~Xu, ``Pfb-diff: Progressive feature blending diffusion for text-driven image editing,'' \emph{Neural Networks}, 2025.

\bibitem{wang2023instructedit}
Q.~Wang, B.~Zhang, M.~Birsak, and P.~Wonka, ``Instructedit: Improving automatic masks for diffusion-based image editing with user instructions,'' \emph{arXiv preprint arXiv:2305.18047}, 2023.

\bibitem{tang2024locinv}
C.~Tang, K.~Wang, F.~Yang, and J.~van~de Weijer, ``Locinv: Localization-aware inversion for text-guided image editing,'' \emph{CVPR 2024 AI4CC workshop}, 2024.

\bibitem{wang2024vision}
K.~Wang, X.~Song, M.~Liu, J.~Yuan, and W.~Guan, ``Vision-guided and mask-enhanced adaptive denoising for prompt-based image editing,'' \emph{arXiv preprint arXiv:2410.10496}, 2024.

\bibitem{chefer2023attend}
H.~Chefer, Y.~Alaluf, Y.~Vinker, L.~Wolf, and D.~Cohen-Or, ``Attend-and-excite: Attention-based semantic guidance for text-to-image diffusion models,'' in \emph{SIGGRAPH}, 2023.

\bibitem{rassin2024linguistic}
R.~Rassin, E.~Hirsch, D.~Glickman, S.~Ravfogel, Y.~Goldberg, and G.~Chechik, ``Linguistic binding in diffusion models: Enhancing attribute correspondence through attention map alignment,'' in \emph{NeurIPS}, 2023.

\bibitem{xie2023boxdiff}
J.~Xie, Y.~Li, Y.~Huang, H.~Liu, W.~Zhang, Y.~Zheng, and M.~Z. Shou, ``Boxdiff: Text-to-image synthesis with training-free box-constrained diffusion,'' in \emph{ICCV}, 2023, pp. 7452--7461.

\bibitem{ge2023expressive}
S.~Ge, T.~Park, J.-Y. Zhu, and J.-B. Huang, ``Expressive text-to-image generation with rich text,'' in \emph{ICCV}, 2023.

\bibitem{dahary2024yourself}
O.~Dahary, O.~Patashnik, K.~Aberman, and D.~Cohen-Or, ``Be yourself: Bounded attention for multi-subject text-to-image generation,'' in \emph{ECCV}, 2024.

\bibitem{LiuComposite}
J.~Liu, T.~Huang, and C.~Xu, ``Training-free composite scene generation for layout-to-image synthesis,'' in \emph{ECCV}, 2025.

\bibitem{diffeditor}
C.~Mou, X.~Wang, J.~Song, Y.~Shan, and J.~Zhang, ``Diffeditor: Boosting accuracy and flexibility on diffusion-based image editing,'' in \emph{CVPR}, 2024.

\bibitem{song2020score}
Y.~Song, J.~Sohl-Dickstein, D.~P. Kingma, A.~Kumar, S.~Ermon, and B.~Poole, ``Score-based generative modeling through stochastic differential equations,'' in \emph{ICLR}, 2021.

\bibitem{ho2022classifier}
J.~Ho and T.~Salimans, ``Classifier-free diffusion guidance,'' in \emph{NeruIPS workshop}, 2021.

\bibitem{cho2023noise}
H.~Cho, J.~Lee, S.~B. Kim, T.-H. Oh, and Y.~Jeong, ``Noise map guidance: Inversion with spatial context for real image editing,'' in \emph{ICLR}, 2023.

\bibitem{patashnik2023localizing}
O.~Patashnik, D.~Garibi, I.~Azuri, H.~Averbuch-Elor, and D.~Cohen-Or, ``Localizing object-level shape variations with text-to-image diffusion models,'' in \emph{ICCV}, 2023.

\bibitem{lu2023tf}
S.~Lu, Y.~Liu, and A.~W.-K. Kong, ``Tf-icon: Diffusion-based training-free cross-domain image composition,'' in \emph{ICCV}, 2023.

\bibitem{OpenAI2023GPT4TR}
OpenAI, ``Gpt-4 technical report,'' \emph{arXiv preprint arXiv:2303.08774}, 2023.

\bibitem{facebook2:online}
\BIBentryALTinterwordspacing
A.~Kirillov, E.~Mintun, N.~Ravi, H.~Mao, C.~Rolland, L.~Gustafson, T.~Xiao, S.~Whitehead, A.~C. Berg, W.-Y. Lo, P.~Doll{\'a}r, and R.~Girshick, ``Segmentanything model,'' 2023. [Online]. Available: \url{https://github.com/facebookresearch/segment-anything}
\BIBentrySTDinterwordspacing

\bibitem{CompViss92:online}
\BIBentryALTinterwordspacing
R.~Rombach, A.~Blattmann, D.~Lorenz, P.~Esser, and B.~Ommer, ``High-resolution image synthesis with latent diffusion models,'' 2022. [Online]. Available: \url{https://huggingface.co/CompVis/stable-diffusion-v1-4}
\BIBentrySTDinterwordspacing

\bibitem{omerbtSp27:online}
\BIBentryALTinterwordspacing
N.~Tumanyan, O.~Bar-Tal, S.~Bagon, and T.~Dekel, ``Splicing vit features for semantic appearance transfer,'' 2022. [Online]. Available: \url{https://github.com/omerbt/Splice}
\BIBentrySTDinterwordspacing

\bibitem{showlabl4:online}
\BIBentryALTinterwordspacing
J.~Z. Wu, X.~Li, D.~Gao, Z.~Dong, J.~Bai, A.~Singh, X.~Xiang, Y.~Li, Z.~Huang, Y.~Sun, R.~He, F.~Hu, J.~Hu, H.~Huang, H.~Zhu, X.~Cheng, J.~Tang, M.~Z. Shou, K.~Keutzer, and F.~Iandola, ``Cvpr 2023 text guided video editing competition,'' 2023. [Online]. Available: \url{https://github.com/showlab/loveu-tgve-2023}
\BIBentrySTDinterwordspacing

\bibitem{mturk:online}
\BIBentryALTinterwordspacing
``Amazon mechanical turk.'' [Online]. Available: \url{https://requester.mturk.com/create/projects/new}
\BIBentrySTDinterwordspacing

\bibitem{p2p:online}
\BIBentryALTinterwordspacing
A.~Hertz, R.~Mokady, J.~Tenenbaum, K.~Aberman, Y.~Pritch, and D.~Cohen-Or, ``Prompt-to-prompt image editing with cross attention control,'' 2022. [Online]. Available: \url{https://github.com/google/prompt-to-prompt}
\BIBentrySTDinterwordspacing

\bibitem{pnp:online}
\BIBentryALTinterwordspacing
N.~Tumanyan, M.~Geyer, S.~Bagon, and T.~Dekel, ``Plug-and-play diffusion features for text-driven image-to-image translation,'' 2023. [Online]. Available: \url{https://github.com/MichalGeyer/plug-and-play}
\BIBentrySTDinterwordspacing

\bibitem{spdinv:online}
\BIBentryALTinterwordspacing
R.~Li, R.~Li, S.~Guo, and L.~Zhang, ``Source prompt disentangled inversion for boosting image editability with diffusion models,'' 2024. [Online]. Available: \url{https://github.com/leeruibin/SPDInv}
\BIBentrySTDinterwordspacing

\bibitem{guide_rescale:online}
\BIBentryALTinterwordspacing
V.~Titov, M.~Khalmatova, A.~Ivanova, D.~Vetrov, and A.~Alanov, ``Guide-and-rescale: Self-guidance mechanism for effective tuning-free real image editing,'' 2024. [Online]. Available: \url{https://github.com/AIRI-Institute/Guide-and-Rescale}
\BIBentrySTDinterwordspacing

\bibitem{magedit:online}
\BIBentryALTinterwordspacing
Q.~Mao, L.~Chen, Y.~Gu, Z.~Fang, and M.~Z. Shou, ``Mag-edit: Localized image editing in complex scenarios via mask-based attention-adjusted guidance,'' 2024. [Online]. Available: \url{https://github.com/HelenMao/MAG-Edit}
\BIBentrySTDinterwordspacing

\bibitem{iedit:online}
\BIBentryALTinterwordspacing
Q.~Wang, B.~Zhang, M.~Birsak, and P.~Wonka, ``Instructedit: Improving automatic masks for diffusion-based image editing with user instructions.'' 2023. [Online]. Available: \url{https://github.com/QianWangX/InstructEdit}
\BIBentrySTDinterwordspacing

\bibitem{qi2023fatezero}
C.~Qi, X.~Cun, Y.~Zhang, C.~Lei, X.~Wang, Y.~Shan, and Q.~Chen, ``Fatezero: Fusing attentions for zero-shot text-based video editing,'' in \emph{ICCV}, 2023.

\end{thebibliography}

\end{document}